\documentclass[11pt,a4paper]{article}
\usepackage{times,latexsym}
\usepackage{url}
\usepackage[T1]{fontenc}

\usepackage{microtype}

\usepackage{inconsolata}

\usepackage{graphicx}
\usepackage{booktabs}
\usepackage{amsmath}
\usepackage{amsfonts}
\usepackage{diagbox}
\usepackage{multirow}
\usepackage{siunitx}
\usepackage{float}
\usepackage{textcomp}
\usepackage{subcaption}
\usepackage{etoolbox}
\restylefloat{table}

\usepackage[acceptedWithA]{tacl2021v1}

\usepackage{xspace,mfirstuc,tabulary}

\newif\iftaclinstructions
\taclinstructionsfalse %
\iftaclinstructions
\renewcommand{\confidential}{}
\renewcommand{\anonsubtext}{(No author info supplied here, for consistency with
TACL-submission anonymization requirements)}
\newcommand{\instr}
\fi

\iftaclpubformat %

\else

\fi

\title{Explicitly Representing Syntax Improves Sentence-to-layout Prediction of Unexpected Situations}

\author{
  Wolf Nuyts\Thanks{Joint first authors.} \quad
  Ruben Cartuyvels\footnotemark[1] \ \Thanks{Corresponding author.} \quad
  Marie-Francine Moens%
  \\
  Department of Computer Science
  \\
  KU Leuven, Belgium
  \\
    \texttt{wolf.nuyts@gmail.com, \ \{ruben.cartuyvels, \ sien.moens\}@kuleuven.be}
}

\date{}

\newcommand{\Repw}{\text{Re}_{Dpw}}
\newcommand{\Prpw}{\text{Pr}_{Dpw}}
\newcommand{\Fpw}{\text{F1}_{Dpw}}

\newcommand{\Rer}{\text{Re}^\text{repl}}

\newcommand{\Fp}{\text{F1}_{0.5}}
\newcommand{\Rp}{\text{Re}_{0.5}}
\newcommand{\Pp}{\text{Pr}_{0.5}}

\newtoggle{showold}
\togglefalse{showold}

\begin{document}
\maketitle
\begin{abstract}

Recognizing visual entities in a natural language sentence and arranging them in a 2D spatial layout require a compositional understanding of language and space. This task of layout prediction is valuable in text-to-image synthesis as it allows localized and controlled in-painting of the image. In this comparative study it is shown that we can predict layouts from language representations that implicitly or explicitly encode sentence syntax, if the sentences mention similar entity-relationships to the ones seen during training.
To test compositional understanding, we collect a test set of grammatically correct sentences and layouts describing compositions of entities and relations that unlikely have been seen during training. Performance on this test set substantially drops, showing that current models rely on correlations in the training data and have difficulties in understanding the structure of the input sentences. 
We propose a novel structural loss function that better enforces the syntactic structure of the input sentence and show large performance gains in the task of 2D spatial layout prediction conditioned on text. 
The loss has the potential to be used in other generation tasks where a tree-like structure underlies the conditioning modality.
Code, trained models and the USCOCO evaluation set will be made available via github\footnote{\url{https://github.com/rubencart/USCOCO}}.

\iftoggle{showold}{
Recognizing visual entities in a natural language sentence and arranging them in a 2D spatial layout requires a compositional understanding of language and space. 
We show that neural networks are able to predict layouts from language representations computed by popular transformer based language models that are said to implicitly capture syntax, if the sentences mention similar entity relationships to the ones seen during training.
To test compositional understanding, we have collected a test set of grammatically correct sentences and layouts describing compositions of entities and relations that are unlikely to occur in the training data.
Performance drastically drops on this test set, showing that the current models 
rely on correlations in the training data and have difficulties understanding the grammar in input sentences. 
\textcolor{red}{Explicitly integrating %
syntax structure 
into %
semantic representations does not improve compositional generalization by itself, but we show large performance gains when the syntax structure is additionally enforced by a new loss that we propose.}
}{}

\end{abstract}

\section{Introduction}
Current neural networks and especially transformer architectures pretrained on large amounts of data build powerful representations of content. However unlike humans, they fail when confronted with unexpected situations and content which is out of context \cite{Geirhos2020}. Compositionality is considered as a powerful tool in human cognition as it enables humans to understand and generate a potentially infinite number of novel situations by viewing the situation as a novel composition of familiar simpler parts \cite{humboldt1999language,chomsky1965aspects,frankland2020compositionality}.

\begin{figure}[ht!]
\begin{center}
\includegraphics[trim={0 5pt 0 5pt},clip,width=0.45\textwidth]{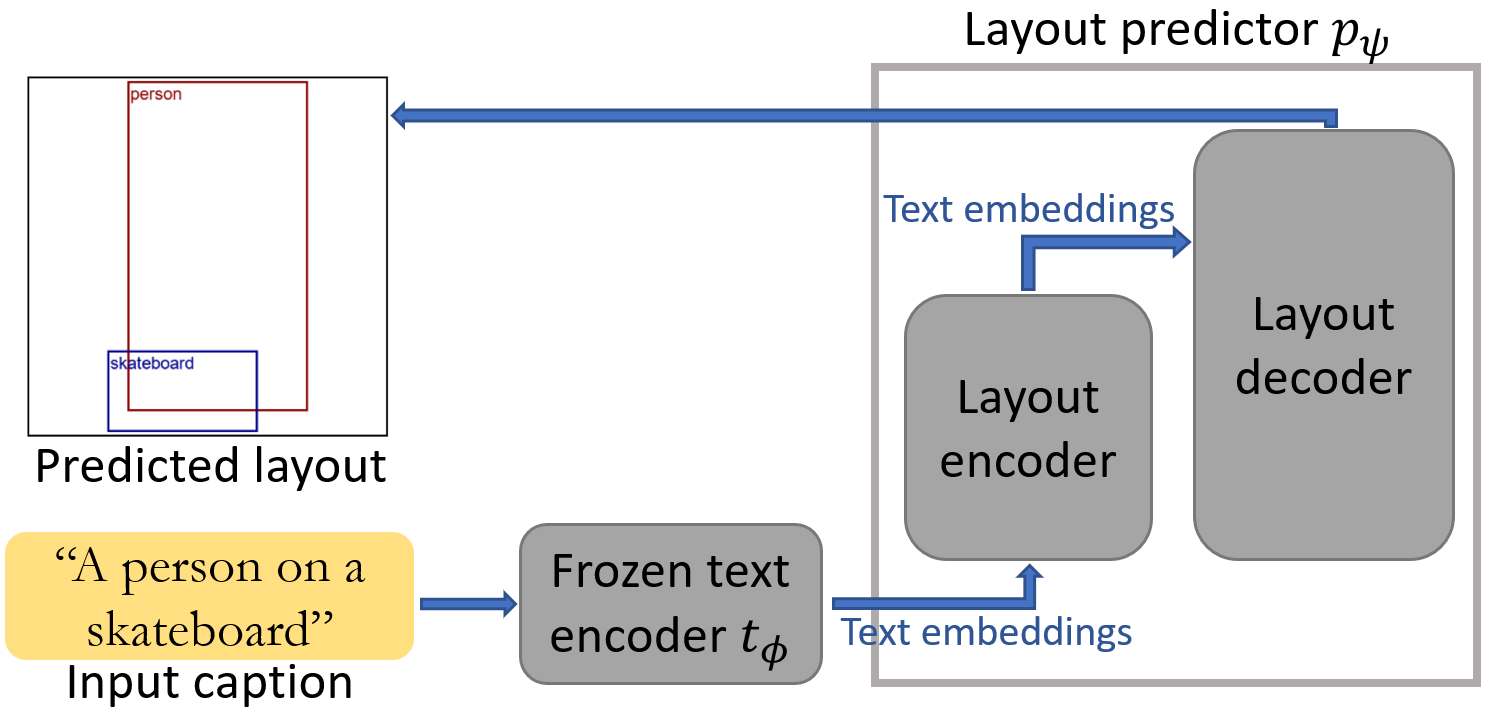}
\caption{\small Overview of text-to-layout prediction.}
\label{fig:model_overview}
\end{center}
\end{figure}

In this paper we hypothesize that representations that better encode the syntactical structure of a sentence - in our case a constituency tree - are less sensitive to a decline in performance when confronted with unexpected situations. 
We test this hypothesis with the task of 2D visual object layout prediction given a natural language input sentence (Figure \ref{fig:model_overview} gives an overview of the task and of our models). 
We collect a test set of grammatically correct sentences and layouts (visual ``imagined'' situations), called  Unexpected Situations of Common Objects
in Context (USCOCO) describing compositions of entities and relations that unlikely have been seen during training. Most importantly, we propose a novel structural loss function that better retains the syntactic structure of the sentence in its representation by enforcing the alignment between the syntax tree embeddings and the output of the downstream task, in our case the visual embeddings. 
This loss function is evaluated both with models that explicitly integrate syntax (i.e., linearized constituent trees using brackets and tags as tokens) and with models that implicitly encode syntax (i.e., language models trained with a transformer architecture). 
Models that explicitly integrate syntax
show large performance gains in the task of 2D spatial layout prediction conditioned on text
when using the proposed structural loss.

\begin{figure}[ht!]
\begin{center}
\includegraphics[clip,trim={0 0 0 0},width=0.45\textwidth]{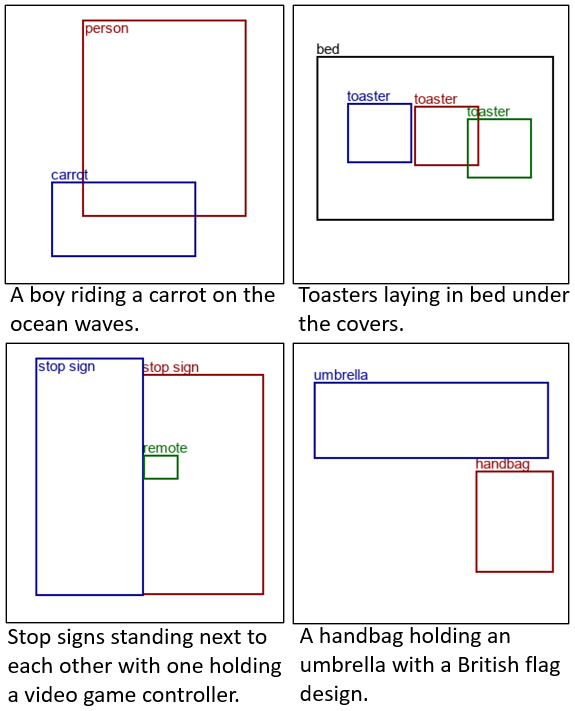}
\caption{\small Samples from the USCOCO dataset.}
\label{fig:absurd_examples}
\end{center}
\end{figure}

The task of layout prediction is valuable in text-to-image synthesis as it allows localized and controlled in-painting of the image. 
Apart from measuring the understanding of natural language by the machine \cite{Ulinski2019}, text-to-image synthesis is popular because of the large application potential (e.g., when creating games and virtual worlds). Current generative diffusion models create a naturalistic image from a description in natural language \citep[e.g., DALL-E 2,][]{ramesh2022dalle2}, but lack local control in the image triggered by the role a word has in the interpretation of the sentence \cite{rassin2022dalle2}, and 
fail to adhere to specified relations between objects \citep{gokhale2022visor}.
Additionally, if you change the description, for instance, by changing an object name or its attribute, a new image is generated from scratch, instead of a locally changed version of the current image, which is a concern in research \citep[e.g.,][]{couairon2022diffedit,poole2022dreamfusion}.
\citet{chen2023control, qu2023layoutllmt2i} show that using scene layouts as additional input greatly improves the spatial reasoning of text-to-image models, substantiating the importance of the text-to-layout synthesis task. 
We restrict the visual scene to the spatial 2D arrangements of objects mentioned in the natural language sentence, taking into account the size and positions of the objects (Figure \ref{fig:absurd_examples}).
From these layouts images can be generated that accurately adhere to the spatial restrictions encoded by the layouts \cite{chen2023control, qu2023layoutllmt2i}, but this is not within the scope of this paper. 
We emphasize that while we argue that good layout predictions are valuable, the question this study aims to answer is not how to build the best possible layout predictors, but whether and if so how explicitly representing syntax can improve such predictors, especially with respect to their robustness to unseen and unexpected inputs.

The contributions of our research are the following: 1) We introduce a new test set called USCOCO for the task of 2D visual layout prediction based on a natural language sentence containing unusual interactions between known objects; 
2) We compare multiple sentence encoding models that implicitly and explicitly integrate syntactical structure in the neural encoding, and evaluate them with the downstream task of layout prediction; 
3) We introduce a novel parallel decoder for layout prediction based on transformers; 
4) We propose a novel contrastive structural loss that enforces the encoding of syntax in the representation of a visual scene and show that it increases generalization to unexpected compositions; 
5) We conduct a comprehensive set of experiments, using quantitative metrics, human evaluation and probing to evaluate the encoding of structure, more specifically syntax.

\iftoggle{showold}{%
Current neural networks and especially transformer architectures pretrained on large amounts of data build powerful representations of content. However, they fail when confronted with unexpected situations and content which is 
out of context \cite{Geirhos2020}. Humans can perfectly deal with such a situation. In this paper we hypothesize that representations that capture the syntactical structure of language utterances - in our case a constituency tree - and that retain its compositional property are less sensitive to a decline in performance. 
{Compositionality is considered as a powerful tool in human cognition as it enables humans to understand and generate a potentially infinite number of novel situations by viewing the situation as a novel composition of familiar simpler parts \cite{humboldt1999language,chomsky1965aspects,frankland2020compositionality}.}
In this paper we evaluate this hypothesis in the specific task of visualizing the content of a sentence through an image, as the translation of language to a visual modality is one way to evaluate the understanding of language by the machine \cite{Ulinski2019}. 
\textcolor{blue}{Moreover, there is a current interest in text-to-image synthesis which has led to the development of generative diffusion models that create a naturalistic image from a description in natural language (e.g., DALL-E 2). It is known that DALL-E 2 ignores that each word has a single role in the interpretation \cite{rassin2022dalle2} and sometimes fails to correctly bind a word in a syntactic relationship. Additionally, if you change the description, for instance, by changing an object name or its attribute, a complete new image is generated. Current research investigates how to locally change the image (e.g., \cite{couairon2022diffedit,poole2022dreamfusion}). Such a functionality requires a compositional modeling and is very valuable if a user, for instance, creates a virtual world in a gaming environment or creates advertisements. The layout prediction task by which we evaluate the embedding of syntax in sentence representations contributes to the goal of creating compositional representations that are valuable in text-to-image sythesis.}  
{Layouts to guide stable-diffusion to model spatial relations mentioned in text better \cite{chen2023control}. 
Evaluation benchmark for image generation models that uses captions with spatial relations between random COCO categories, then detects whether correctly modeled in output (very similar to our absurd test set idea!) \cite{gokhale2022visor}.}
We create a novel dataset of grammatically well formed sentences of unusual and sometimes absurd events and their visual ``imagined'' situations, called Unexpected Situations of Common Objects in Context (USCOCO). We restrict the visual scene to the spatial 2D arrangements of the objects mentioned in the natural language sentence taking into account the size and positions of the objects (Figure \ref{fig:absurd_examples}). From these layouts the pixels of the objects possibly refined by their pose could be predicted \cite{li2019objectdriven,sylvain2020layouts,chen2023control}, but this is not the scope of this paper. 
}{}

\iftoggle{showold}{\begin{figure}[ht!]
\begin{center}
\includegraphics[trim={0 0pt 0 3pt},clip,width=0.45\textwidth]{figures/indom_samples.png}
\caption{Conventional samples from the in-domain dataset.}
\label{fig:conv_examples}
\end{center}
\end{figure}}{}

\section{Related work}

\textbf{Implicit syntax in language and visio-linguistic models}
Deep learning has ruled out the need for feature engineering in natural language processing including the extraction of syntactical features. With the advent of contextualized language models pretrained on large text corpora \cite{Peters2018, devlin2019bert} representations of words and sentences are dynamically computed. Several studies have evaluated syntactical knowledge embedded in language models through targeted syntactic evaluation, probing, and downstream natural language understanding tasks \cite{hewitt2019probe, Manning2020,Linzen2021,Kulmizev2021}. 
Here we consider the task of visually imagining the situation expressed in a sentence and show that in case of unexpected situations that are unlikely to occur in the training data 
the performance of language models strongly decreases. 

\textbf{Probing 
compositional multimodal reasoning
}
One of the findings of our work, that is, the failure of current models in sentence-to-layout prediction of unusual situations, are in line with recent text-image alignment test suites like Winoground \cite{thrush2022winoground} and VALSE 
\cite{parcalabescu2022valse}, that ask to correctly retrieve an image using grammatically varying captions.
The benchmark of \citet{gokhale2022visor} considers a generation setting, like us, however, they evaluate image generation (down to pixels), while %
ignoring the role of the language encoder\footnote{Their data has not been made publicly available at the time of writing.}.
Moreover, their captions are automatically generated from object names and simple spatial relations, and hence only contain explicit spatial relations, whereas our USCOCO captions are modified human-written COCO captions and include implicit spatial relations.

\textbf{Explicitly embedding syntax in 
language representations}.
\citet{Popa2021} use a tensor factorisation model for computing token embeddings that integrate dependency structures. In section \ref{sec:explicit-syntax} we discuss models that integrate a constituency parse tree \cite{qian2021structural, sartran2022transformergram} as we use them as encoders in the sentence-to-layout prediction task. 
They build on %
generative parsing approaches using recurrent neural networks \citep{dyer2016rnngs,choe2016parsing}.

\textbf{Visual scene layout prediction}
\citet{hong2018inferring}, \citet{Tan2019} and \citet{li2019layoutgan} introduce %
layout predictors that are similar to our autoregressive model. They also train on the realistic COCO dataset and generate a dynamic number of objects, but they do not investigate the layout predictions.
Other layout prediction methods 
require structured input like triplets or graphs instead of free-form text \citep{li2019sg2sl,johnson2018sg2im,lee2020design}, are confined to layouts of only 2 objects \citep{collell2021probing}, are unconditional \citep{li2019layoutgan}, work in simplified, non-realistic settings \citep{radevski2020decoding,lee2020design} or focus on predicting positions for known objects and their relationships \citep{radevski2020decoding}.

\section{Methods}
\subsection{Task definition}

Given a natural language caption $C$, the task is to generate a layout $L$ that captures a spatial 2D visual arrangement of the objects that the caption describes.
A layout $L = \{ b_i \}_i = \{(o_i, x_i, y_i, w_i, h_i) \}_i$ consists of a varying number of visual objects, each represented by a 5-tuple, where $o_i$ is a label from a category vocabulary (in this paper: one of the 80 COCO categories, e.g., ``\textit{elephant}''), and where $(x_i, y_i, w_i, h_i)$ refers to the bounding box for that object. The coordinates of the middle point of a box are $x_i, y_i$, and $w_i, h_i$ are the width and height.
A caption $C$ consists of a number $N_c$ of word tokens $c_i$.
Hence, we want to learn the parameters $\theta$ of a model $f_\theta$ that maps captions to layouts: $L = f_\theta( C )$.

\textbf{Model overview}
We split the prediction problem into two parts, as shown in Figure \ref{fig:model_overview}. First, a text encoder $t_\phi$ computes (potentially structured) text embeddings $e_j$ for input word tokens $c_i$: $E = t_\phi(C)$. 
Second, a layout predictor $p_\psi$ predicts an embedding $v_k$ per visual object: $v_k = p_\psi(E)$. These are projected by a multilayer perceptron to a categorical probability distribution to predict object labels: $o_k \sim \text{softmax}(\text{MLP}^\text{label}(v_k))$. Regression
($\times 4$) is used for positions $b_k = (x_k, y_k, w_k, h_k)$: e.g., $x_k = \sigma(\text{MLP}^\text{pos}(v_k)_x)$, with $\sigma: \mathbb{R}^d \mapsto [0, 1]$ a sigmoid.

\subsection{Text encoders $t_\phi$}

We consider two types of text encoders. First, we %
explicitly encode the syntactical structure of a sentence in its semantic representation. As syntactical structure we choose a constituency parse as it naturally represents the structure of human language, and phrases can be mapped to visual objects. 
Human language is characterized by recursive structures which correspond with the recursion that humans perceive in the world \cite{Hawkings2021, Hauser2002recursion}. 
Second, we encode the sentence with a state-of-the-art sentence encoder that is pretrained with a
next token prediction objective. We assume that it implicitly encodes syntax \citep{tenney2019probing,warstadt2020blimp}. The choice of the next-token prediction objective is motivated by existing 
work that compares language models with implicit vs. explicit syntax \citep{qian2021structural,sartran2022transformergram}.
We use pretrained text encoders, which are all frozen during layout prediction training to allow a clean comparison.
 
\subsubsection{Explicitly embedding syntax in sentence representations}
\label{sec:explicit-syntax}

The models that explicitly embed syntax take a linearized version of the %
constituency trees as input, using brackets and constituent tags as tokens in addition to the input sentence $C$, to end up with $C^\text{lin}$. 
For instance, ``\textit{A dog catches a frisbee}'' would be preprocessed into ``(NP \textit{a dog}) (VP \textit{catches} (NP \textit{a frisbee}))''. 
The constituency trees are obtained with the parser of \citet{kitaev2018constituency,kitaev2019multilingual}. The model computes an embedding $e^{\text{struct}}_j \ (0 \leq j < N_c + N_\text{lin})$ per input token $c_j$ (including the parentheses and syntax tags). 
The embeddings are given as is to the layout predictor, and since tokens explicitly representing structure (parentheses, syntax tags) have their own embedding, we assume the sequence of $e^{\text{struct}}_j$ to carry more structural information than the sequence formed by $e^{\text{base}}_i$ (cfr. section \ref{sec:te-baselines}).
We consider the following models, which are all pretrained on the BLLIP\textsubscript{LG} dataset \citep[$\approx$40M tokens]{charniak2000bllip}.

\textbf{PLM}: The Parsing as Language Model from \citet{qian2021structural} inputs $C^\text{lin}$ into an untrained GPT-2 model and learns $e^{\text{struct}}_j$ by training on a next token prediction task.

\textbf{PLM}\textsubscript{mask}: %
of \citet{qian2021structural} is similar to \texttt{PLM} but uses masking to constrain two of the attention heads in the transformer layers to attend to tokens that are respectively part of the current constituent and part of the rest of the partially parsed sentence. 

\textbf{TG}: The Transformer Grammars model from \citet{sartran2022transformergram} uses a masking scheme that 
constrains all attention heads to only attend to local parts of the constituent tree.
This results in a recursive composition of the representations of smaller constituents
into the representations of larger constituents, which reflects closely the recursive property of Recurrent Neural Network Grammars (RNNG) models \citep{dyer2016rnngs}.
We adapt this model to use a GPT-2 backbone and we train it for next token prediction on the same dataset as the \texttt{PLM} models for a fair comparison. 

{\textbf{TG}\textsubscript{RB}: To test to what extent differences in layout prediction performance are due to the explicit use of a constituency grammar, and not a byproduct of model and/or input differences, this model uses trivial right-branching constituency trees (constructed by taking the silver trees and moving all closing brackets to the end of the sentence), instead of silver constituency trees, both during pretraining and during layout generation. This model is also used by \cite{sartran2022transformergram} as ablation baseline.
}

\subsubsection{Baselines that are assumed to implicitly encode syntax}
\label{sec:te-baselines}

The baselines take a sequence of text tokens $c_i \ (0 \leq i < N_c)$ and produce a sequence of the same length, of embeddings $e^{\text{base}}_i \ (0 \leq i < N_c)$, which will be given to the layout decoder.

\textbf{GPT-2}\textsubscript{Bllip}: %
    This language model is also used by  \citet{qian2021structural} and shares its architecture and training regime with 
    GPT-2 \citep{radford2019language}. 
    It is trained on the sentences (not the linearized parse trees) of the BLLIP\textsubscript{LG} dataset to predict the next token given the history \citep{charniak2000bllip}. 
    Hence, this model is trained on the same sentences as the models of section \ref{sec:explicit-syntax}.
    {Even though it is debatable whether transformers can learn implicit syntax from the relatively small ($\approx$40M tokens) BLLIP\textsubscript{LG} dataset, this model is used as baseline by existing work on explicit syntax in language modelling \citep{qian2021structural,sartran2022transformergram}, which is why we also include it.}
    {
    Furthermore, there is evidence that pretraining datasets of 10M-100M tokens suffice for transformers to learn most of their syntax capabilities  \cite{perez2021howmuch,zhang2021whendo,samuel2023inshape}, even though orders of magnitude ($>$1B) more pretraining tokens are required for more general downstream NLU tasks.}

\textbf{GPT-2}:
As published by \citet{radford2019language}, trained for next token prediction on a large-scale scraped webtext dataset.

$\textbf{GPT-2}_\text{Bllip}^\text{shuffle}$:
Identical to the \texttt{GPT-2}\textsubscript{Bllip} model but the tokens in the input sentence are randomly shuffled%
, to test whether syntax has any contribution at all.
{The pretraining is exactly the same as \textbf{GPT-2}\textsubscript{Bllip}, with the token order preserved.
}

{\textbf{LLaMA}: Large state-of-the-art language models trained on massive amounts of text, we use the 7B and 33B model variants \citep{touvron2023llama}.}

\subsection{Layout predictors $p_\psi$}
\label{sec:layout-predictors}
\subsubsection{Models}
\label{sec:layout-predictor-models}

As a baseline we consider the layout prediction LSTM from \citet{hong2018inferring} and \citet{li2019objectdriven}, further referred to as \texttt{ObjLSTM}\footnote{This is the only existing model we found that generates varying numbers of bounding boxes from free-form text for the COCO dataset.}.
This model does not perform well and it trains slowly because of the LSTM architecture, so we propose two 
novel layout predictors.
The two models use the same transformer encoder, and differ in their transformer decoder architecture.

    \textbf{{PAR}}: This decoding model is inspired by the DETR model for object detection of \citet{carion2020detr}. The decoder generates a number of objects in a single forward parallel pass, after first predicting the number of objects.

    \textbf{SEQ}: This autoregressive model is similar to the language generating transformer of \citet{vaswani2017attention}, but decodes object labels and bounding boxes and not language tokens. It predicts an object per step until the end-of-sequence token is predicted, or the maximum length is reached. The model is similar to the layout decoder of \citet{li2019objectdriven}, but uses transformers instead of LSTMs.

\subsubsection{Training}
\label{sec:layout-predictor-training}

\paragraph{The \texttt{PAR} model} is trained analogous to \citet{carion2020detr} and \citet{stewart2016peopledet} by first computing the minimum cost bipartite matching between predicted objects $\hat{b}_i$ and ground-truth objects $b_j$ (as the ordering might differ), using the Hungarian algorithm \cite{kuhn2010hungarian}, with differences in box labels, positions and overlaps as cost. 

\paragraph{The \texttt{SEQ} model} is trained to predict the next visual object given the previous GT objects.
The order of generation is imposed (by a heuristic: from large to small in area, after \citet{li2019objectdriven}). 
The 1\textsuperscript{st}, 2\textsuperscript{nd},... generated object is matched with the largest, 2\textsuperscript{nd} but largest,... ground-truth object.

The \texttt{PAR} and \texttt{SEQ} models apply the following losses to 
each matched pair of predicted box $\hat{b}_i$ and ground-truth box $b_j$.

    \paragraph{A cross-entropy loss} $\mathcal{L}_{\text{label}}(\hat{b}_i, {b}_j)$ applied to the object labels.
    \paragraph{A combination of regression losses} applied to the bounding box coordinates. 
        \begin{itemize}
            \item An L1-loss $\mathcal{L}_{\text{L1}}(\hat{b}_i, {b}_j)$ applied to each of the dimensions of the boxes 
        \cite{carion2020detr}.
            \item The generalized box IoU loss $\mathcal{L}_{\text{gIoU}}(\hat{b}_i, {b}_j)$ proposed by \citet{rezatofighi2019giou}, taking into account overlap of boxes.
            \item An L1-loss $\mathcal{L}_{\text{prop}}$ taking into account the proportion of box width and height.
            \iftoggle{showold}{{:
              $\mathcal{L}_{\text{prop}}(\hat{b}_i, {b}_j) = |\frac{w_{i}-h_{i}}{w_{i}+h_{i}} - \frac{\hat{w}_{j}-\hat{h}_{j}}{\hat{w}_{j}+\hat{h}_{j}} | \nonumber$.}}{}
            \item A loss $\mathcal{L}_{\text{rel}}$ equal to the difference between predicted and ground-truth predictions of relative distances between objects. 
            \iftoggle{showold}{{Distance $D$ denotes Euclidean distance between middle points, scaled by the diagonal lengths of the boxes.
            $(\hat{b}_k, {b}_m)$ denotes another matched pair of bounding boxes in the same sample: $\mathcal{L}_{\text{rel}}(\hat{b}_i, {b}_j) = \sum_{(k,m)} |{D}(b_{j}, b_{m}) - {D}(\hat{b}_i, \hat{b}_k) |$.
            }}{}
        \end{itemize}
The following losses are not applied between matched predicted and ground-truth object pairs, but to the entire sequence of output objects at once.

    \paragraph{A cross-entropy loss} $\mathcal{L}_{\text{len}}$ to the predicted number of object queries (\texttt{PAR} only).
    
\paragraph{A contrastive structural loss} %
    to enforce in a novel way the grammatical structure found in the parse trees
    on the output, in our case the visual object embeddings $v_k$ that are computed by the layout predictor and that are used to predict the object boxes and their labels. 
    
    To calculate the loss, all nodes in the parse tree, that is, leaf nodes corresponding to word tokens, and parent and root nodes corresponding to spans of word tokens, are represented separately by a positional embedding $e_j^\text{pos} (0 \leq j < 2N_c -1)$ following \citet{shiv2019novel}. 
    The positional embeddings are learnt, {they are agnostic of the content of the word or word spans they correspond to,} and they encode the path through the tree, starting from the root, to the given node.
    
    In a contrastive manner the loss forces the visual object representations $v_k$
    to be close to the positional embeddings $e^\text{pos}_j$,
    but far from those $\hat{e}^\text{pos}_j$ of other sentences in the minibatch.
    It maximizes the posterior probability that the set of visual object embeddings $V_m = \{v_k\}_k$ for sample $m$ are matched with the set of tree positional embeddings $E_m = \{e_j^\text{pos}\}_j$ for the same sample $m$,
    and vice versa:
    $\mathcal{L}_\text{struct}(m) = - \log P(E_m|V_m) - \log P(V_m|E_m)$.
    These probabilities are computed as a softmax over similarity scores $S(m,n)$ between samples in the batch, where the denominator of the softmax sums over tree positions or objects, resp.
    
    The similarity score for 2 samples $m,n$ is computed as a log-sum-exp function of the cosine similarities between the $i$'th visual object $v_i^m$ of sample $m$ and a visually informed syntax tree context vector $c^n_i$ representing all tree positions $\{e^n_j\}_j$ of sample $n$.
    The context vector $c^n_i$ is computed with the attention mechanism of \citet{bahdanau2015align}, with tree positional embeddings $\{e^n_j\}_j$ 
    as keys and values, and visual embedding $v^m_i$ 
    as query.
    The dot product between query $v^m_i$ and keys $\{e^n_j\}_j$ is first additionally normalized over the visual objects corresponding to a tree position, before the regular normalization over tree positions.
    {These normalized dot products between keys and queries constitute a soft matching between visual objects and constituency tree node positions (note: only the positions, representing syntax and not semantics of the words, are represented by the tree positional embeddings). Since the model learns this mapping from the training signal provided by this loss, it is not necessary to manually specify which text spans are to be matched to which visual objects.}
    Appendix \ref{app:struct-loss} contains a complete definition.
    
    The loss has resemblance to the loss used by \citet{xu2018attngan}, replacing their text embeddings by our visual object embeddings, and their visual embeddings by our syntax tree embeddings. 
    {Note that only the constituency parse of the input text and the output embeddings are needed. 
    In this case, the output embeddings represent visual objects, but they are in general not confined to only represent visual objects, they could technically represent anything.
    Hence, the loss is not tied to layout generation in specific, and could be applied to any generation task conditioned on (grammatically) structured text, as tree positions are matched to output embeddings. This novel loss is completely independent of the text encoder and can be applied to a text encoder with explicit syntax input, or to a text encoder with implicit syntax (if a constituency parse of the input is available).\footnote{
    The loss uses explicit syntax in the form of a constituency parse, so when used to train a model with implicit syntax as input (like \texttt{GPT-2}\textsubscript{Bllip}, which does not use linearized parse trees $C^\text{lin}$ as input), it adds explicit syntax information to the training signal.
    Nevertheless, in this study, we call such a model an ``implicit syntax model with structural loss''
    }
    }
    
\paragraph{The final loss} for one training sample is the sum of the above losses 
$\mathcal{L}$,
with $(\hat{b}_i, {b}_j)$ matched pairs of predicted and GT objects, and each loss weighted by a different weight $\lambda$.
\begin{align}
    \mathcal{L}_\text{total} &=  \lambda_\text{1} \mathcal{L}_{\text{struct}} + \lambda_\text{2} \mathcal{L}_\text{len} \nonumber \\
    &+ \sum_{(i,j)} \Big[ \lambda_\text{3} \mathcal{L}_{\text{label}}(\hat{b}_i, {b}_j) + \lambda_\text{4} \mathcal{L}_{\text{L1}}(\hat{b}_i, {b}_j) \nonumber \\
    &+ \lambda_\text{5} \mathcal{L}_{\text{gIoU}}(\hat{b}_i, {b}_j) + \lambda_\text{6} \mathcal{L}_{\text{prop}}(\hat{b}_i, {b}_j)  \label{eq:total-loss} \nonumber \\
    &+ \lambda_\text{7} \mathcal{L}_{\text{rel}}(\hat{b}_i, {b}_j) \Big]
\end{align}

\subsection{Datasets}
\label{sec:datasets}

\begin{table}
    \centering
    \footnotesize
    \begin{tabular*}{\linewidth}{@{\extracolsep{\fill}} lrr }
    \toprule
     $t_\phi$ & Train set & Regime   \\
    \midrule
    \texttt{PLM} & BLLIP sents, trees & NTP \\
    \texttt{PLM}\textsubscript{mask} & BLLIP sents, trees & NTP \\
    \texttt{TG} & BLLIP sents, trees & NTP \\
    \texttt{GPT-2}\textsubscript{Bllip} & BLLIP sents & NTP \\
    \texttt{GPT-2} & $\approx$8B text tokens &  NTP \\ %
    \texttt{LLaMA} & 1.4T tokens & NTP \\ %
    \bottomrule
    \end{tabular*}
    \caption{Overview of the text encoders,
    their training data and training regimes. NTP stands for next-token prediction.
    }
    \label{tab:model-overview}
\end{table}

The text encoders are pretrained on datasets summarized in Table \ref{tab:model-overview}.
We use COCO captions 
and instances \citep[bounding boxes and labels,][]{lin2014microsoft} for training and testing the layout decoder.
We use the 2017 COCO split with 118K training images and 5K validation images (both with 5 captions per image). The testing images are not usable as they have no captions and bounding box annotations. 
We randomly pick 5K images from the training data for validation and use the remaining 113K as training set $\mathcal{D}_\text{train}$. 
We use the 2017 COCO validation set as in-domain test set $\mathcal{D}_\text{indom}$. 
$\mathcal{D}_\text{USCOCO}$ is our 
test set of unexpected situations with 2.4K layouts and 1 caption per layout.

\paragraph{Collection of USCOCO}
We used Amazon Mechanical Turk (AMT) to collect ground-truth (caption, layout)-pairs denoting situations that are unlikely to occur in the training data. We obtained this test set in three steps.
In the \textbf{first} step, we asked annotators to link sentence parts of captions in $\mathcal{D}_\text{indom}$ to bounding boxes.

\textbf{Second}, we used a script to replace linked sentence parts in the captions with a random COCO category name ($o_\text{new}$, with a different COCO supercategory than the bounding box had that the sentence part was linked to).
The script also replaces the bounding box that the annotators linked to the replaced sentence part in the first step, with a bounding box for an object of the sampled category $o_\text{new}$. 
We use 4 replacement strategies: the first keeps the original box merely replacing its label. 
The next 3 strategies also adjust the size of the box based on average dimensions of boxes with category $o_\text{new}$ in $\mathcal{D}_\text{train}$, relative to the size of the nearest other box in the layout.
The 2\textsuperscript{nd} places the middle point of the new box on the middle point of the replaced box, the 3\textsuperscript{rd} at an equal x-distance to the nearest object box and the the 4\textsuperscript{th} at an equal y-distance to the nearest object.

In the \textbf{third} step, annotators were shown the caption with the automatically replaced sentence part and the 4 corresponding automatically generated layouts.
They were asked to evaluate whether the new caption is grammatically correct, and which of the 4 layouts fits the caption best (or none).
Each sample of step 2 was verified by 3 different annotators. Samples where at least 2 annotators agreed on the same layout and none of the 3 annotators considered the sentence as grammatically incorrect, 
were added to the final USCOCO dataset.
Appendix \ref{app:uscoco} contains a more complete description of the collection process.

The USCOCO test set 
follows a very different distribution of object categories than $\mathcal{D}_\text{train}$. 
To show this we calculate co-occurrences of object categories in all images (weighted so that every image has an equal impact) of $\mathcal{D}_\text{train}$, $\mathcal{D}_\text{USCOCO}$ and $\mathcal{D}_\text{indom}$.
The co-occurence vectors of $\mathcal{D}_\text{train}$ and $\mathcal{D}_\text{USCOCO}$ have a cosine similarity of $47\%$, versus $99\%$ for $\mathcal{D}_\text{train}$ and the in-domain test set $\mathcal{D}_\text{indom}$.

\subsection{Preprocessing of the images} \label{sec:preprocessing}

\paragraph{Spurious bounding boxes (\texttt{SP})}
Because 
objects annotated with bounding boxes in the COCO images are not always mentioned in the corresponding captions, we implement a filter for bounding boxes and apply it on all train and test data.
The filter computes for each object class $O$ of COCO the average diagonal length $\bar{d}_O$ of its bounding box, over the training set, and the normalized average diagonal length $\bar{d}^\text{norm}_O$ (scaled by the size of the biggest object of each image). 
Only the biggest object of a class per image is included in these averages to limit the influence of background objects. 
Then, all the objects with size smaller than $0.5 \cdot \bar{d}_O$ and normalized size smaller than $0.5 \cdot \bar{d}^\text{norm}_O$ are discarded. %
\iftoggle{showold}{{
We use $L_O$ as shorthand for all boxes in layout $L$ of category $O$, if $L_O$ is empty, the $\max$-operator returns 0, and $N_O$ denotes the number of images with at least one object of category $O$.
\begin{equation}
    \bar{d}_O =  \frac{1}{N_O}\sum_{L \in \mathcal{D}_\text{train}} \max_{b_i \in L_o} d_i
\end{equation}
\begin{equation}
    \bar{d}^\text{norm}_O =  \frac{1}{N_O}\sum_{L \in \mathcal{D}_\text{train}} \frac{\max_{b_i \in L_o} d_i}{\max_{b_i \in L} d_i}
\end{equation}
}}{}
The normalized threshold allows the filters to be scale invariant, while the non-normalized threshold removes filtering mistakes when there is a big unimportant, unmentioned object in the image. 

\paragraph{Crop-Pad-Normalize (\texttt{CPN})}
To center and scale bounding boxes, we follow %
\citet{collell2021probing}. 
We first crop the tightest enclosing box that contains all object bounding boxes.
Then, we pad symmetrically the smallest side 
to get a square box of height and width $P$.
This preserves the aspect ratio when normalizing. %
Finally, we normalize coordinates by $P$, resulting in coordinates in $[0,1]$.

\subsection{Evaluation metrics} \label{sec:evaluation-metrics}
\paragraph{Pr, Re, F1} precision, recall and F1 score of predicted object labels, without taking their predicted bounding boxes into account.
\paragraph{$\boldsymbol{\Pp}$, $\boldsymbol{\Rp}$, $\boldsymbol{\Fp}$}
precision, recall and F1 score of predicted object labels, with an Intersection over Union (IoU) threshold of 0.5 considering the areas of the predicted and ground-truth bounding boxes %
\citep{ren2017faster}.
The matching set $M_{IoU}$ between ground-truth (GT) and predicted objects is computed in a greedy fashion based on box overlap in terms of pixels.

\paragraph{$\boldsymbol{\Rer}$}
The recall (without positions) on only the set of GT objects that have been replaced in the test set of unexpected situations $\mathcal{D}_{\text{USCOCO}}$.

\begin{table*}[ht!]
    \centering
    \footnotesize
\begin{tabular*}{\linewidth}{@{\extracolsep{\fill}} lllllll }

\toprule
 & \multicolumn{3}{c}{$\mathcal{D}_\text{indom}$} 
 & \multicolumn{3}{c}{$\mathcal{D}_\text{USCOCO}$} 
\\
    {} 
     & \multicolumn{1}{c}{F1$_{0.5}\uparrow$} 
     & \multicolumn{1}{c}{F1 $\uparrow$} 
     & $\Fpw \uparrow$ 
    & \multicolumn{1}{c}{F1$_{0.5}\uparrow$} 
     & \multicolumn{1}{c}{F1 $\uparrow$} 
     & $\Fpw \uparrow$ 
\\
\midrule
\texttt{ObjLSTM}* & .185 {\tiny $\pm$ .021} & \textbf{.676} {\tiny $\pm$ .006} & .356 {\tiny $\pm$ .019} & .099 {\tiny $\pm$ .013} & .524 {\tiny $\pm$ .009} & .16 {\tiny $\pm$ .019} \\
\midrule
\midrule
\texttt{ObjLSTM}\textcolor{white}{\textsubscript{sm}} + \texttt{GPT-2}\textsubscript{Bllip} & .104 {\tiny $\pm$ .003} & .542 {\tiny $\pm$ .01} & .238 {\tiny $\pm$ .013} & .074 {\tiny $\pm$ .003} & .404 {\tiny $\pm$ .014} & .078 {\tiny $\pm$ .007} \\
\texttt{ObjLSTM}\textsubscript{lrg} + \texttt{GPT-2}\textsubscript{Bllip} & .167 {\tiny $\pm$ .005} & .65 {\tiny $\pm$ .006} & .345 {\tiny $\pm$ .01} & .1 {\tiny $\pm$ .003} & .524 {\tiny $\pm$ .016} & .174 {\tiny $\pm$ .019} \\
\texttt{SEQ} + \texttt{GPT-2}\textsubscript{Bllip} & .271 {\tiny $\pm$ .004} & .597 {\tiny $\pm$ .01} & .304 {\tiny $\pm$ .011} & .167 {\tiny $\pm$ .001} & .485 {\tiny $\pm$ .006} & .149 {\tiny $\pm$ .007} \\
\texttt{PAR} + \texttt{GPT-2}\textsubscript{Bllip} & \textbf{.296} {\tiny $\pm$ .004} & {.67} {\tiny $\pm$ .014} & \textbf{.375} {\tiny $\pm$ .018} & \textbf{.18} {\tiny $\pm$ .001} & \textbf{.576} {\tiny $\pm$ .026} & \textbf{.229} {\tiny $\pm$ .036} \\
\midrule
\texttt{SEQ} + \texttt{TG} & .28 {\tiny $\pm$ .002} & .638 {\tiny $\pm$ .006} & .344 {\tiny $\pm$ .011} & .177 {\tiny $\pm$ .002} & .541 {\tiny $\pm$ .002} & .203 {\tiny $\pm$ .004} \\
\texttt{PAR} + \texttt{TG} & \textbf{.306} {\tiny $\pm$ .008} & \textbf{.69} {\tiny $\pm$ .002} & \textbf{.398} {\tiny $\pm$ .008} & \textbf{.185} {\tiny $\pm$ .004} & \textbf{.6} {\tiny $\pm$ .004} & \textbf{.255} {\tiny $\pm$ .005} \\
\bottomrule
\end{tabular*}
    
    \caption{
    \texttt{PAR}, \texttt{SEQ} and \texttt{ObjLSTM} (baseline) layout predictor results on USCOCO and $\mathcal{D}_\text{indom}$, incl. F1, $\Fp$ and $\Fpw$ showing that the PAR decoding model performs best.
    All entries use the \texttt{GPT-2}\textsubscript{Bllip} 
    or \texttt{TG} 
    text encoder (without structural loss), except for \texttt{ObjLSTM}* which uses a multimodal text encoder trained on images and text \citep{li2019objectdriven,xu2018attngan}. 
    \texttt{ObjLSTM}\textsubscript{lrg} is scaled up to same number of parameters as \texttt{SEQ} and \texttt{PAR}, and uses a layout predictor with a transformer encoder before the LSTM decoder, like \texttt{SEQ} and \texttt{PAR}.
    Best results of models using implicit syntax (upper rows) and those with explicit syntax (lower rows, with \texttt{TG})
    are marked in bold.
    }
    \label{tab:results-par}
\end{table*}

\paragraph{$\boldsymbol{\Prpw}$, $\boldsymbol{\Repw}$, $\boldsymbol{\Fpw}$}

The $\Fp$ score penalizes an incorrect/missing label as much as it penalizes an incorrect position, while we consider an incorrect/missing label to be a worse error. 
Additionally, there are many plausible spatial arrangements for one caption (as explained in section \ref{sec:preprocessing} image preprocessing  tries to reduce its impact).
For this reason we introduce an F1 score based on the precision and recall of object pairs, penalized by the difference of the distance between the two boxes in the GT and the two boxes in the predictions.
This metric penalizes incorrect positions, since a pair's precision or recall gets downweighted when its distance is different from its distance in the GT, but it penalizes incorrect labels more, since pairs with incorrect labels have precision/recall equal to 0. 
Moreover, it evaluates positions of boxes {relative} to each other, instead of to one absolute GT layout.

First, a greedy matching set $M_D$ between GT and predicted objects is computed based on labels and middle-point distance.
Boxes $\hat{b}$ are part of the predicted layout $L_\text{p}$ (with $N_\text{p}$ object pairs), and boxes $b$ are part of the GT layout $L_\text{GT}$ (with $N_\text{GT}$ object pairs).
The matching function $\pi(\hat{b}_{i}, \hat{b}_{k})$ equals 1 if predicted boxes $\hat{b}_{i}$ and $\hat{b}_{k}$ both have a matching box $b_{j}$ and $b_{m}$ in the GT (so $(i,j) \in M_D$ and $(k,m) \in M_D$), and equals 0 otherwise.
$D$ denotes Euclidean distance between box middle points, and $S_{jm,ik} \in [0, 1]$ is a normalized similarity metric based on this distance.
The penalized precision $\Prpw$ and recall $\Repw$ are computed as follows:
\begin{align}
    S_{jm,ik} &= 1 - \frac{1}{\sqrt{2}} |{D}(b_{j}, b_{m}) - {D}(\hat{b}_{i}, \hat{b}_{k}) | \nonumber  \\
    \Prpw &= \frac{1}{N_\text{p}} \sum_{i\in L_\text{p}} \sum_{k \in L_\text{p} \setminus \{i\}} S_{jm,ik} \, \pi(\hat{b}_{i}, \hat{b}_{k}) \\
    \Repw &= \frac{1}{N_\text{GT}} \sum_{i\in L_\text{p}} \sum_{k \in L_\text{p} \setminus \{i\}} S_{jm,ik} \, \pi(\hat{b}_{i}, \hat{b}_{k}) \\
    \text{where } &(i, j) \in M_D \ \text{and} \ (k, m) \in M_D . \nonumber
\end{align}
${\Fpw}$ is finally computed as the standard F1 of ${\Prpw}$ and ${\Repw}$.
If a sample has less than 2 boxes in the GT or predictions, respectively ${\Repw}$ or ${\Prpw}$ is undefined for that sample.\footnote{
There are less than 2 boxes for 0\% of samples in $\mathcal{D}_\text{USCOCO}$ and 19\% in $\mathcal{D}_\text{indom}$.
Re and $\Rp$ are defined for samples with only 1 box, and samples with 0 boxes do almost not occur.
}

\section{Experiments}

\subsection{Experimental set-up}
All runs were repeated three times and the averages and standard deviations are reported. 
We used a learning rate of $10^{-4}$ with Adam \citep{kingma2014adam}, a batch size of 128 (64 for runs using $\mathcal{L}_\text{struct}$), random horizontal flips on the bounding boxes as data augmentation, and early stopping.
All text encoders were frozen.
Layout predictors use a hidden dimension of 256\footnote{
We use the same dimension regardless of the text encoder to allow for a fair comparison. Increasing the dimension did not improve results.
} and a FFN dimension of 1024, with 4 encoder layers and 6 decoder layers, and have 10M parameters.
The loss weights (eq. \ref{eq:total-loss}) were chosen experimentally and set to $\lambda_\text{1}\in\{0.25, 0.5, 1.0\}$, $\lambda_\text{2}=0.1$, $\lambda_\text{3}=0.5$, $\lambda_\text{4}=5$, $\lambda_\text{5}=2$, $\lambda_\text{6}=0.5$, $\lambda_\text{7}=0.5$.
We took most of the other \texttt{PAR} hyperparameters from \citet{carion2020detr}.
{
We run all text encoders with the smallest {GPT-2} architecture (125M params), for which we reuse checkpoints shared by \citet{qian2021structural} for \texttt{PLM}, \texttt{PLM}\textsubscript{mask} and \texttt{GPT-2}\textsubscript{Bllip}.}
{We also run \texttt{GPT-2-lg}\textsubscript{Bllip}, \texttt{GPT-2-lg} and \texttt{TG-lg} with the larger GPT-2 architecture (755M params).
\texttt{GPT-2} and \texttt{GPT-2-lg} runs use checkpoints from Huggingface \cite{wolf2020huggingface}, and \texttt{LLaMA} runs use checkpoints shared by Meta.
We train 
\texttt{GPT-2-lg}\textsubscript{Bllip} ourselves, using the code of \citet{qian2021structural}.
}
Models were trained on one 16GB Tesla P100 or 32GB Tesla V100 GPU (except the \texttt{LLaMA-33B} runs which were trained on a 80GB A100).

\paragraph{Training \texttt{TG}} 
We train \texttt{TG} and \texttt{TG-lg} like \texttt{PLM} and baseline \texttt{GPT-2}\textsubscript{Bllip} following \citet{qian2021structural}, with a learning rate $10^{-5}$, the AdamW optimizer, a batch size of 5 and trained until convergence on the development set of the {BLLIP}\textsubscript{LG} dataset split \citep{charniak2000bllip}.
We implement \texttt{TG} with the recursive masking procedure of \citet{sartran2022transformergram}, but without the relative positional encodings, since these do not contribute much to the performance, and because \texttt{GPT-2} uses absolute position embeddings.

\begin{table*}[ht!]
    \footnotesize
\centering
\begin{tabular*}{\linewidth}{@{\extracolsep{\fill}} llllllllllll }
\toprule
 {} & & & \multicolumn{3}{c}{$\mathcal{D}_\text{USCOCO}$} \\
  & & Size %
     & $\Fpw \uparrow$ 
     & $\Fp\uparrow$
     & F1 $\uparrow$
     \\
\midrule

\texttt{GPT-2} & & 125M & .207 {\tiny $\pm$ .019} & .179 {\tiny $\pm$ .008} & .566 {\tiny $\pm$ .02}  \\
\texttt{GPT-2} & + $\mathcal{L}_\text{struct}$ & & .187 {\tiny $\pm$ .006} & .184 {\tiny $\pm$ .007} & .555 {\tiny $\pm$ .011}  \\

\texttt{GPT-2}\textsubscript{Bllip} & & & .229 {\tiny $\pm$ .036} & .18 {\tiny $\pm$ .001} & .576 {\tiny $\pm$ .026}  \\
\texttt{GPT-2}\textsubscript{Bllip} & + $\mathcal{L}_\text{struct}$ &  & .233 {\tiny $\pm$ .014} & .192 {\tiny $\pm$ .003} & .574 {\tiny $\pm$ .014} \\
\midrule

\texttt{GPT-2-lg} & & 755M & .283 {\tiny $\pm$ .047} & .188 {\tiny $\pm$ .005} & .61 {\tiny $\pm$ .03}  \\
\texttt{GPT-2-lg} & + $\mathcal{L}_\text{struct}$ & & .292 {\tiny $\pm$ .025} & .205 {\tiny $\pm$ .009} & .628 {\tiny $\pm$ .016} \\
\texttt{GPT-2-lg}\textsubscript{Bllip} & & & .233 {\tiny $\pm$ .027}  & .183 {\tiny $\pm$ .002}  & .586 {\tiny $\pm$ .019} \\
\texttt{GPT-2-lg}\textsubscript{Bllip} & + $\mathcal{L}_\text{struct}$ & & .234 {\tiny $\pm$ .019} & .196 {\tiny $\pm$ .005} & .579 {\tiny $\pm$ .006}  \\ \midrule
\texttt{LLaMA-7B} & & 7B & .231 {\tiny $\pm$ .014} & .179 {\tiny $\pm$ .007}  & .583 {\tiny $\pm$ .011}  \\
\texttt{LLaMA-7B} & + $\mathcal{L}_\text{struct}$ & & .26 {\tiny $\pm$ .026} & .192 {\tiny $\pm$ .01}  & .602 {\tiny $\pm$ .02}  \\

\midrule
\midrule

\texttt{PLM} & & 125M & .226 {\tiny $\pm$ .006} & .18 {\tiny $\pm$ .002} & .579 {\tiny $\pm$ .002} \\
\texttt{PLM} & + $\mathcal{L}_\text{struct}$ & & .282 {\tiny $\pm$ .048} & .192 {\tiny $\pm$ .002}  & .61 {\tiny $\pm$ .033}  \\

\texttt{PLM}\textsubscript{mask} & & & .234 {\tiny $\pm$ .012} & .176 {\tiny $\pm$ .005} & .588 {\tiny $\pm$ .01} \\
\texttt{PLM}\textsubscript{mask} & + $\mathcal{L}_\text{struct}$ & & .28 {\tiny $\pm$ .039} & .191 {\tiny $\pm$ .007} & .612 {\tiny $\pm$ .024} \\

\texttt{TG} & & & .255 {\tiny $\pm$ .005} & .185 {\tiny $\pm$ .004} & .6 {\tiny $\pm$ .004} \\
\texttt{TG} & + $\mathcal{L}_\text{struct}$ & & .318 {\tiny $\pm$ .026} & .192 {\tiny $\pm$ .008} & .641 {\tiny $\pm$ .018} \\
\midrule
\texttt{TG-lg} & & 755M & .283 {\tiny $\pm$ .017} & .183 {\tiny $\pm$ .008} & .621 {\tiny $\pm$ .014}  \\
\texttt{TG-lg} & + $\mathcal{L}_\text{struct}$ & & .327 {\tiny $\pm$ .018} & .195 {\tiny $\pm$ .006} & .645 {\tiny $\pm$ .01} \\

\bottomrule
\end{tabular*}

    \caption{
    Text encoders with implicit (above double line)
    and explicit (below double line)
    syntax, and structural loss results on USCOCO: F1, precision and recall, with and without IoU threshold and pairwise distance weighted. All entries use the \texttt{PAR} layout predictor. 
    Results of the best (in terms of $\Fpw$) $\lambda_\text{1}$ for each model type are shown.
    }
    \label{tab:results-qian}

\end{table*}

\section{Results and discussion}

\subsection{Layout prediction}

\subsubsection{Preprocessing of images}

{We ran a comparison of preprocessing for the \texttt{PLM} and \texttt{GPT-2}\textsubscript{Bllip} text encoders (both using \texttt{PAR}). All conclusions were identical.}
{
Using \texttt{SP} gives small but significant improvements in F1$_{0.5}$ and F1 on both test sets,
and larger improvements when also normalizing bounding boxes with \texttt{CPN}.}
Using \texttt{CPN} increases the position sensitive F1$_{0.5}$ metric drastically on both test sets, even more so when also using \texttt{SP}.
In a human evaluation with AMT, annotators chose the best layout given a COCO caption from $\mathcal{D}_\text{indom}$. 
500 captions with 2 corresponding layouts (one from $\mathcal{D}_\text{indom}$ + \texttt{CPN} + \texttt{SP} and one from $\mathcal{D}_\text{indom}$ + \texttt{CPN}) were evaluated by 3 annotators,
who preferred \texttt{SP} in 37\% of cases, as opposed to 18.6\% where they preferred not using \texttt{SP}
(44.4\% of the time they were indifferent).
These results suggest that the preprocessing techniques improve the alignment of COCO bounding boxes with their captions, and that the best alignment is achieved when using both.

\subsubsection{Layout prediction models}

Table \ref{tab:results-par} compares our new \texttt{PAR} and \texttt{SEQ} layout predictors with the \texttt{ObjLSTM} baseline.
All models use either the \texttt{GPT-2}\textsubscript{Bllip} or \texttt{TG} text encoder (based on the small GPT-2 architecture), except for \texttt{ObjLSTM}* which uses a multimodal %
text encoder following \citet{li2019objectdriven,xu2018attngan}.
The $\mathcal{L}_\text{rel}$ and $\mathcal{L}_\text{prop}$ losses are used for the \texttt{SEQ} and \texttt{PAR} runs (in Table \ref{tab:results-par} and subsequent Tables).
These losses give minor consistent improvements in $\Fp$, while keeping F1 more or less constant.

Both \texttt{SEQ} / \texttt{PAR} + \texttt{GPT-2}\textsubscript{Bllip} models outperform all \texttt{ObjLSTM} baselines by significant margin on the position sensitive $\Fp$ metric on both test sets (even though \texttt{ObjLSTM}* uses a text encoder that has been pretrained on multimodal data).
\texttt{PAR} + \texttt{GPT-2}\textsubscript{Bllip} obtains better $\Fpw$ and position insensitive F1 scores than baselines on the unexpected test set, and similar $\Fpw$ and F1 on $\mathcal{D}_\text{indom}$.
\texttt{SEQ} + \texttt{GPT-2}\textsubscript{Bllip} lags a bit behind on the last 2 metrics.

\texttt{PAR} obtains a significantly better precision than \texttt{SEQ}, both with and without object positions (Pr and $\Pp$), on both test sets, both with \texttt{GPT-2}\textsubscript{Bllip} and \texttt{TG}, resulting in greater F1 scores.
This could be attributed to the fact that the $n$\textsuperscript{th} prediction with \texttt{SEQ} is conditioned only on the text and $n-1$ preceding objects, while with \texttt{PAR}, all predictions are conditioned on the text and on all other objects.
The fact that for generating language, autoregressive models like \texttt{SEQ} are superior to non-autoregressive models like \texttt{PAR}, but vice versa for generating a set of visual objects, may be due to the inherent sequential character of language, as opposed to the set of visual objects in a layout, which does not follow a natural sequential order.
When generating a set of objects in parallel, the transformer's self-attention can model all pairwise relationships between objects before assigning any positions or labels.
In contrast, when modelling a sequence of objects autoregressively, the model is forced to decide on the first object's label and position without being able to take into account the rest of the generated objects, and it cannot change those decisions later on.

Since the \texttt{PAR} model scores higher and is more efficient (it decodes all $\hat{b}_i$ in one forward pass, compared to one forward pass per $\hat{b}_i$ for \texttt{SEQ}), we use \texttt{PAR} in subsequent experiments.

\subsection{Improved generalization to USCOCO data with syntax}

\subsubsection{Explicitly modelling syntax}

Table \ref{tab:results-qian} shows layout prediction F1$_{0.5}$, F1 and $\Fpw$ on the USCOCO test set of \texttt{PAR} with implicitly structured \texttt{GPT-2}\textsubscript{Bllip}, \texttt{GPT-2} and \texttt{LLaMA-7B} text encoders (upper half) vs. with explicitly structured \texttt{PLM}, \texttt{PLM}\textsubscript{mask} and \texttt{TG} text encoders (bottom half), with (rows with + $\mathcal{L}_\text{struct}$) and without ($\lambda_\text{1}=0$) structural loss.

Without structural loss, all smaller 125M models achieve very similar F1$_{0.5}$ scores compared to the baseline \texttt{GPT-2}\textsubscript{Bllip},
and only \texttt{TG} is able to slightly 
improve the F1 and $\Fpw$ scores. 
We assume that models with explicit syntax, i.e., that integrate syntax in the input sentence, do not learn to fully utilize the compositionality of the syntax with current learning objectives.

{We observe a noticeable increase over all metrics by using \texttt{GPT-2-lg} compared to \texttt{GPT-2} which is to be expected, while \texttt{GPT-2-lg}\textsubscript{Bllip} and \texttt{GPT-2}\textsubscript{Bllip} perform equally where we assume that the training with a relatively small dataset does not fully exploit the capabilities of a larger model. 
\texttt{TG-lg} obtains similar scores as \texttt{TG} with a small increase for the $\Fpw$ and performs on par with \texttt{GPT-2-lg} while trained on only a fraction of the data.
}

Notable is that the very large \texttt{LLaMA-7B} model performs on par with \texttt{GPT-2}\textsubscript{Bllip} and \texttt{GPT-2-lg}\textsubscript{Bllip}.
{A possible explanation could be the quite drastic downscaling of the 4096-dimensional features of \texttt{LLaMA-7B} to 256 dimensions for our layout predictor by a linear layer. Using a 4096-dimensional hidden dimension for the layout predictor did not improve results. This increased the number of trainable parameters by two orders of magnitude, and the large resulting model possibly overfitted the COCO train set.}

{\texttt{TG} outperforms \texttt{PLM} and \texttt{PLM}\textsubscript{mask} in F1 and $\Fpw$, which proves that restricting the attention masking scheme to follow a recursive pattern according to the recursion of syntax in the sentence helps generalizing to unexpected situations.
This is in line with \citet{sartran2022transformergram} who find that \texttt{TG} text encoders show better syntactic generalization.
}

\subsubsection{Structural loss}
\label{sec:struct-loss}

Table \ref{tab:results-qian} displays in the rows with + $\mathcal{L}_\text{struct}$ the impact of training with our structural loss function, with the best weight $\lambda_\text{1}$ in the total loss in eq. \eqref{eq:total-loss} chosen from $\{0.25, 0.5, 1.0\}$.
{
$\Fp$ and $\Pp$ slightly increase for all models and all $\lambda_\text{1}$ values. $\Rp$ is little affected, except for some explicit syntax models and \texttt{LLaMA} that see a slight increase.}

\paragraph{F1, $\boldsymbol{\Fpw}$} 
{For implicit syntax models, with increasing $\lambda_\text{1}$, Re and $\Repw$ decrease severely.
Pr and $\Prpw$ first increase and then decrease again, resulting in sometimes stable but eventually decreasing F1 scores. 
For models with explicit syntax, Re and $\Repw$ increase most for small loss weights ($\lambda_\text{1}=0.25$), while Pr and $\Prpw$ top at $\lambda_\text{1}=0.5$ or $1.0$. 
Together this causes an improvement in F1 but mainly a sharp rise in $\Fpw$.
These trends are more prominent for small models, but persist for large models, incl. \texttt{LLaMA}.}
$\Fpw$ peaks for \texttt{TG}/\texttt{TG-lg} with $0.25 \cdot \mathcal{L}_\text{struct}$ at 0.318/0.327, which is a $\approx$40\% increase of the baselines' performance of \texttt{GPT-2}\textsubscript{Bllip}/\texttt{GPT-2-lg}\textsubscript{Bllip} at 0.229/0.233.

The loss, enforcing explicit constituency tree structure in the output visual embeddings, trains the layout predictor to not lose the explicit structure encoded by \texttt{TG}, \texttt{PLM} and \texttt{PLM}\textsubscript{mask}. 
This compositional structure causes a disentangled, recursive representation of visual scenes \cite{Hawkings2021,Hauser2002recursion}, facilitating the replacement of objects with unexpected different objects for input sentences 
that contain unexpected combinations of objects.
For models with implicit syntax, the loss tries to enforce a structure that is not explicitly available in the models' input (as opposed to for the models with explicit syntax), which may lead to a more difficult learning objective.

\begin{figure}
\begin{center}
\includegraphics[width=0.49\textwidth]{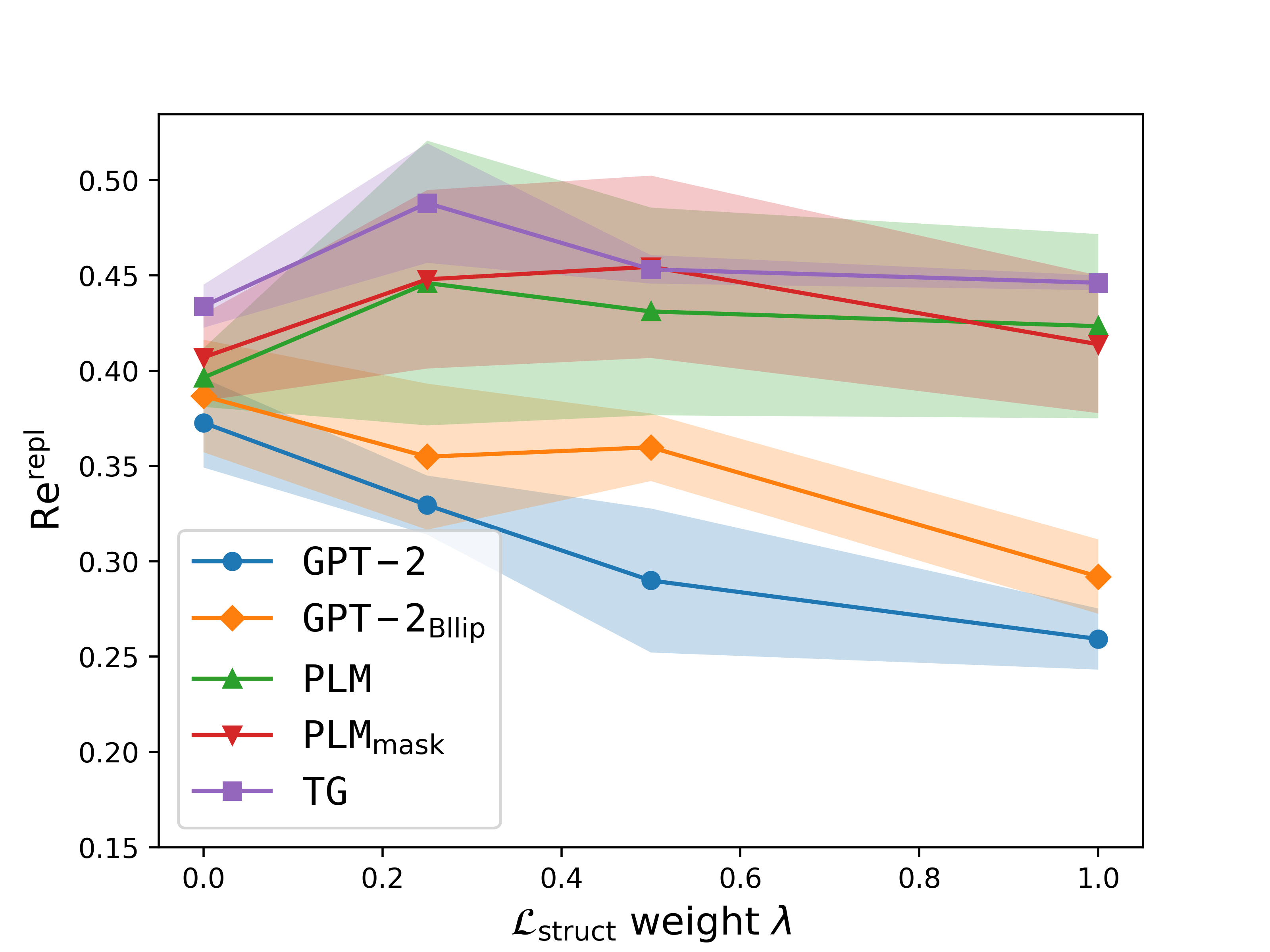}
\caption{\small Recall on replaced objects $\Rer$ in USCOCO vs. structural loss $\mathcal{L}_\text{struct}$ weight $\lambda_\text{1}$.
}
\label{fig:recall_repl}
\end{center}
\end{figure}

\paragraph{$\Rer$} Figure \ref{fig:recall_repl} shows the recall $\Rer$ on the replaced object of USCOCO (the unusual object).
$\Rer$ increases for models with explicit syntax, topping at $\lambda_\text{1}=0.25$, while it decreases sharply for models with implicit syntax \texttt{GPT-2} and $\texttt{GPT-2}_\text{Bllip}$.
\iftoggle{showold}{This confirms that models with explicit syntax get better in predicting unexpected objects with a loss that enforces syntax structure in visual embeddings, while this same loss makes models with implicit syntax significantly weaker at the task.}{}

\begin{table*}[ht!]
    \centering
    \footnotesize
    \begin{tabular*}{\linewidth}{@{\extracolsep{\fill}} lll lll l lll }
    \toprule
    & & &
    \multicolumn{3}{c}{$\mathcal{D}_\text{indom}$} && \multicolumn{3}{c}{$\mathcal{D}_\text{USCOCO}$} \\
   & & Size
     & $\Fp\uparrow$ & F1 $\uparrow$
     & $\Fpw \uparrow$
     && $\Fp\uparrow$ & F1 $\uparrow$
    & $\Fpw \uparrow$ \\
    \midrule
$\texttt{GPT-2}_\text{Bllip}^\text{shuffle}$ & & 125M & .286 {\tiny $\pm$ .006} & .656 {\tiny $\pm$ .005} & .349 {\tiny $\pm$ .012} && .166 {\tiny $\pm$ .002} & .566 {\tiny $\pm$ .003} & .213 {\tiny $\pm$ .004} \\
\midrule
\texttt{GPT-2} & & 125M & .294 {\tiny $\pm$ .004} & .66 {\tiny $\pm$ .01} & .353 {\tiny $\pm$ .019} && .179 {\tiny $\pm$ .008} & .566 {\tiny $\pm$ .02} & .207 {\tiny $\pm$ .019} \\
\texttt{GPT-2}\textsubscript{Bllip} & & & .296 {\tiny $\pm$ .004} & .67 {\tiny $\pm$ .014} & .375 {\tiny $\pm$ .018} & & .18 {\tiny $\pm$ .001} & .576 {\tiny $\pm$ .026} & .229 {\tiny $\pm$ .036} \\
\midrule

\texttt{GPT-2-lg} & & 755M & .308 {\tiny $\pm$ .001} & .702 {\tiny $\pm$ .007} & .414 {\tiny $\pm$ .013} && .188 {\tiny $\pm$ .005} & .61 {\tiny $\pm$ .03} & .283 {\tiny $\pm$ .047} \\
\texttt{GPT-2-lg}\textsubscript{Bllip} & & & .298 {\tiny $\pm$ .004} & .676 {\tiny $\pm$ .01} & .38 {\tiny $\pm$ .013} && .183 {\tiny $\pm$ .002} & .586 {\tiny $\pm$ .019} & .233 {\tiny $\pm$ .027} \\
\midrule
\texttt{LLaMA-7B} & & 7B & .306 {\tiny $\pm$ .001} & .701 {\tiny $\pm$ .003} & .411 {\tiny $\pm$ .008} && .179 {\tiny $\pm$ .007} & .583 {\tiny $\pm$ .011} & .231 {\tiny $\pm$ .014} \\
\texttt{LLaMA-33B} & & 33B & .305 {\tiny $\pm$ .005} & .699 {\tiny $\pm$ .003} & .406 {\tiny $\pm$ .002} && .181 {\tiny $\pm$ .006} & .577 {\tiny $\pm$ .008} & .225 {\tiny $\pm$ .011} \\
\midrule
\midrule
\texttt{TG}\textsubscript{RB} & & 125M & .299 {\tiny $\pm$ .005} & .683 {\tiny $\pm$ .01} & .391 {\tiny $\pm$ .015} && .178 {\tiny $\pm$ .004} & .571 {\tiny $\pm$ .011} & .216 {\tiny $\pm$ .016} \\
\texttt{TG}\textsubscript{RB} & + $\mathcal{L}_\text{struct}$ & & .3 {\tiny $\pm$ .005} & .67 {\tiny $\pm$ .01} & .358 {\tiny $\pm$ .012} && .189 {\tiny $\pm$ .007} & .606 {\tiny $\pm$ .017} & .278 {\tiny $\pm$ .02} \\ 
\midrule
\texttt{PLM} & + $\mathcal{L}_\text{struct}$ & 125M & .301 {\tiny $\pm$ .006} & .677 {\tiny $\pm$ .022} & .378 {\tiny $\pm$ .038} && .192 {\tiny $\pm$ .002} & .61 {\tiny $\pm$ .033} & .282 {\tiny $\pm$ .048} \\
\texttt{PLM}\textsubscript{mask} & + $\mathcal{L}_\text{struct}$ & & .3 {\tiny $\pm$ .004} & .683 {\tiny $\pm$ .003} & .388 {\tiny $\pm$ .007} && .191 {\tiny $\pm$ .007} & .612 {\tiny $\pm$ .024} & .28 {\tiny $\pm$ .039} \\
\texttt{TG} & + $\mathcal{L}_\text{struct}$ & & .305 {\tiny $\pm$ .005} & .685 {\tiny $\pm$ .012} & .379 {\tiny $\pm$ .028} && .192 {\tiny $\pm$ .008} & .641 {\tiny $\pm$ .018} & .318 {\tiny $\pm$ .026} \\
\midrule
\texttt{TG-lg} & + $\mathcal{L}_\text{struct}$ & 755M & .306 {\tiny $\pm$ .004} & .692 {\tiny $\pm$ .002} & .392 {\tiny $\pm$ .007} && .195 {\tiny $\pm$ .006} & .645 {\tiny $\pm$ .01} & .327 {\tiny $\pm$ .018} \\

    \bottomrule
    \end{tabular*}
    \caption{
    Final model results in terms of F1, $\Fp$, and $\Fpw$ on the in-domain (left) and USCOCO (right) test sets. 
    All entries use the \texttt{PAR} layout predictor, and explicit syntax models use $\lambda_\text{1}=0.25$ or $\lambda_\text{1}=0.50$.
    \texttt{GPT-2(-lg)} and \texttt{LLaMA} have been pretrained on much larger text datasets.
    }
    \label{tab:results-final}
\end{table*}

\subsubsection{Overview of explicit vs. implicit syntax}

Table \ref{tab:results-final} gathers results on both test sets\footnote{
Although CLIP \cite{radford2021learning} has been pretrained on multimodal data, and the other text encoders were not (ignoring for a moment the \texttt{ObjLSTM} baseline), we tested CLIP's sentence embedding but results were poor.
}.
The results for all models, but most notably the implicit syntax models, drop significantly on USCOCO compared to $\mathcal{D}_\text{indom}$, 
confirming 
that current state-of-the-art models {struggle with generating unexpected visual scenes.}

\paragraph{$\mathcal{D}_\text{indom}$} Small models that explicitly model syntax obtain slightly better results than the small baseline models for all metrics. 
Models with implicit syntax might perform well on in-domain test data because they have memorized the common structures in training data.
{The large models that were pretrained on huge text datasets (\texttt{GPT-2-lg}, \texttt{LLaMA-xB}) outperform \texttt{TG-lg} on $\mathcal{D}_\text{indom}$, showing that their pretraining does help for this task, but the drop in USCOCO scores suggests that they might overfit the memorized patterns.}
Situations described in COCO captions are commonly found %
in pretraining data, so that syntax is not needed to predict their visual layouts.
The unexpected USCOCO situations however require the extra compositionality offered by explicit syntax.

\paragraph{USCOCO} We clearly see improvement in results of models that explicitly model syntax,
{showing 
the generalization capabilities needed to perform well on the unseen object combinations of USCOCO, provided it is enforced by a correct learning objective as discussed in section \ref{sec:struct-loss}.}
This increase comes without a decrease in performance on the in-domain test data. 
This is important because it will lead to efficient models for natural language processing that can generalize to examples not seen in the training data, exploiting compositionality.

\paragraph{$\texttt{GPT-2}_\text{Bllip}^\text{shuffle}$}
Another indication that the models that implicitly model syntax do not use the structure of natural language to the same extent, but rather exploit co-occurences in training data, is the fact that
$\texttt{GPT-2}_\text{Bllip}^\text{shuffle}$, which is trained to generate layouts from sentences with shuffled words, achieves only slightly worse results than \texttt{GPT-2}\textsubscript{Bllip}, even on the position sensitive $\Fp$ and $\Fpw$ metrics.

\paragraph{\texttt{TG}\textsubscript{RB}} 
{obtains similar scores to \texttt{TG} on in-domain data with trivial right-branching trees.
On unexpected data, performance drops, proving the importance of syntax especially for generalizing to unexpected data.
The structured loss improves generalization shown by USCOCO results, but not to the same extent as for \texttt{TG}.\footnote{
It is not surprising that performance is partially retained with right-branching trees, since English has a right-branching tendency: the F1 overlap between the trivial constituency trees and the silver-truth trees for COCO validation captions is 0.62.
Further, the constituency tags (e.g. ``NP'', ``PP'') are still included, and $N_\text{lin}$ syntax tokens are added to the $N_c$ caption tokens, granting the model more processing power.}
}

\begin{figure}[t!]
    \centering
    \includegraphics[width=0.48\textwidth]{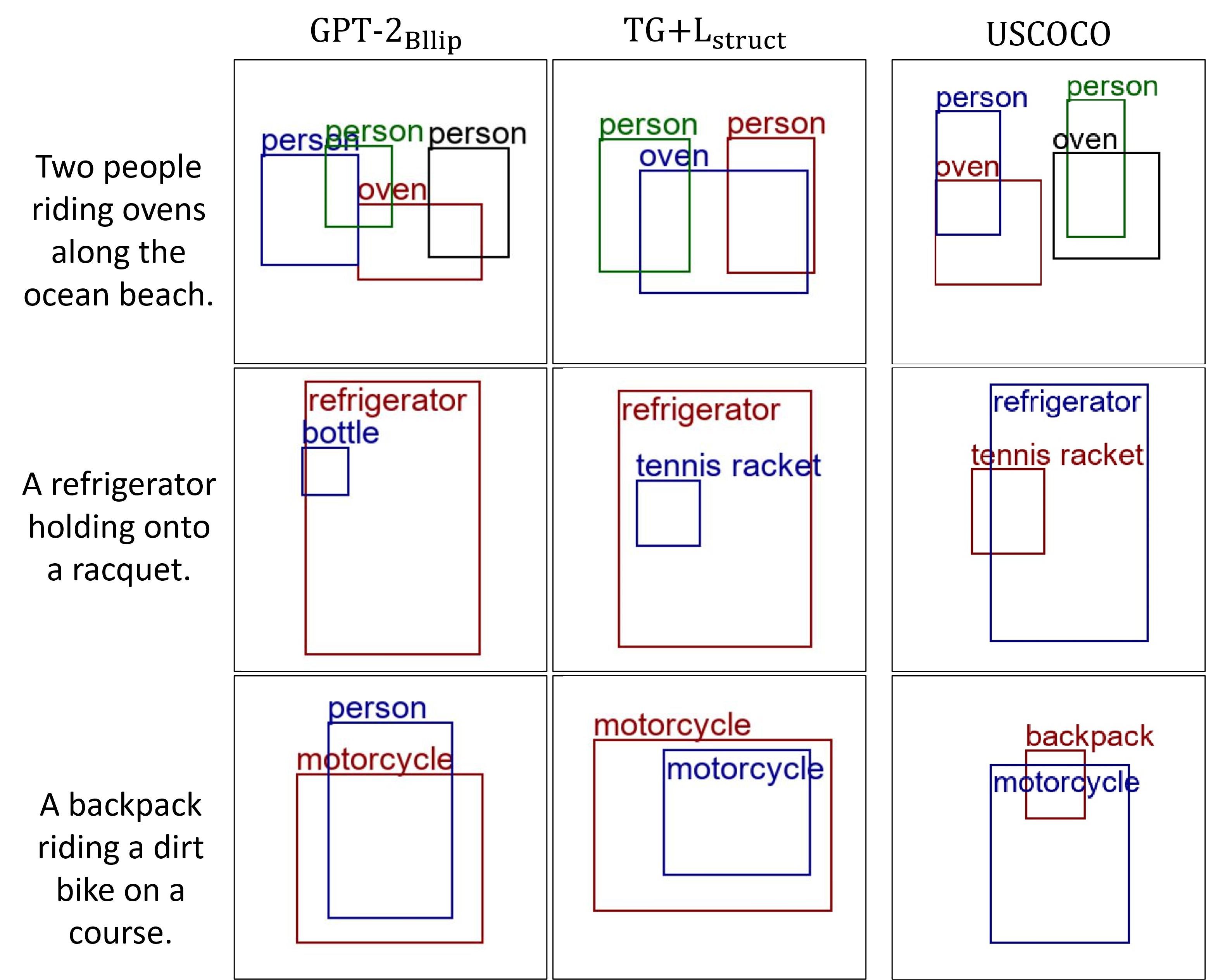}
    \caption{
    \small Examples of generated layouts where annotators chose the layout of \texttt{TG} + $\mathcal{L}_\text{struct}$ over the layout of $\texttt{GPT-2}_\text{Bllip}$ (first 2 examples) and vice versa (last example). }
    \label{fig:qual_results}
\end{figure}

\subsubsection{Human evaluation}
\label{sec:human_eval}

Proper automatic evaluation of performance on the text-to-layout prediction task is hard, since potentially many spatial layouts may fit the scene described by a sentence.
Our metrics compare the predictions to one single ground-truth, ignoring this fact, so we used AMT for a human evaluation of predicted layouts for 500 randomly sampled USCOCO captions. 
For each caption, 3 annotators chose the layout that best fit the caption from a pair of two, based on the following criteria:
Whether the layout displays all objects present in the caption, whether the objects' spatial arrangement corresponds to the caption, whether the objects have reasonable proportions and finally whether object predictions that are not explicitly mentioned in the caption do fit the rest of the scene (i.e., the layout should not contain any absurd or unexpected objects that are not explicitly mentioned in the caption). Figure \ref{fig:qual_results} shows some examples of generated layouts and the annotators' decisions.

The results in Figure \ref{fig:human_eval} are in line with the 
quantitative results where our structural loss proved beneficial for $\texttt{TG}$.
This confirms that explicit structure does not improve layout prediction of unexpected combinations by itself, but together with our structural loss it causes a significant improvement.
We calculated the agreement of the human evaluation with our quantitative metrics,
and found 15.2\% for $\Fp$, 41.7\% for F1 and 42.5\% for $\Fpw$.\footnote{The low percentages were to be expected since metrics often rank layouts equally (when both layouts obtain the same score), while annotators were not given that option.} 
This confirms the previously mentioned suspicion that $\Fp$ is far less suitable for the evaluation of layout generation than F1 and $\Fpw$.

\subsubsection{Constituency tree probes}
\label{sec:probes}
{
To test how the loss affects syntax information in the text embeddings, we run a classifier probe inspired by \citet{tenney2019probing} on the text encoder output and subsequent layers of the encoder of the layout prediction model.
The probe classifies random spans of tokens as being a constituent or not (but ignores their tag).
}

{Figure \ref{fig:probe} shows that all text encoders' outputs (probe layer 0) get good F1 scores and hence do encode syntax.
Without the proposed structural loss, probing results quickly deteriorate in subsequent layers, presumably because the encoder has too little incentive to use and retain constituency structure, because COCO training data contains only situations common in pretraining data and requires no syntactical reasoning.
The figure also shows that it is easier to predict constituency structure from outputs of text encoders with explicit syntax than of those with implicit syntax, which is not surprising because of the former's pretraining and the presence of parentheses and tags.

\begin{figure}[t!]
    \centering
    \includegraphics[clip,trim={0 25pt 0 30pt},width=0.5\textwidth]{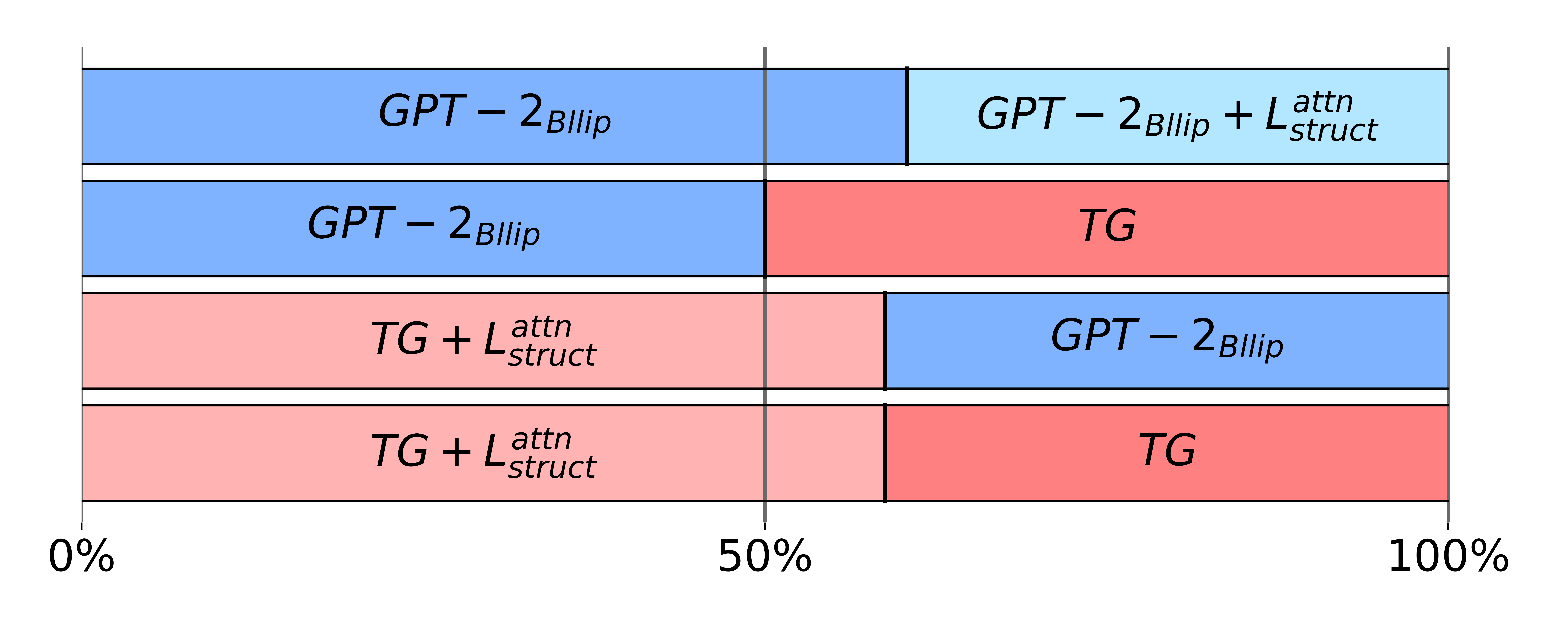}
    \caption{
    \small Human evaluation of generated layouts by \texttt{GPT-2}\textsubscript{Bllip} (+ $\mathcal{L}_\text{struct}$) and \texttt{TG} (+ $\mathcal{L}_\text{struct}$) on USCOCO. Annotators choose the best layout between 2 layouts (anonymized and
order-randomized).}
    \label{fig:human_eval}
\end{figure}

\begin{figure}[t!]
    \centering
    \includegraphics[clip,trim={0 22pt 0 22pt},width=0.48\textwidth]{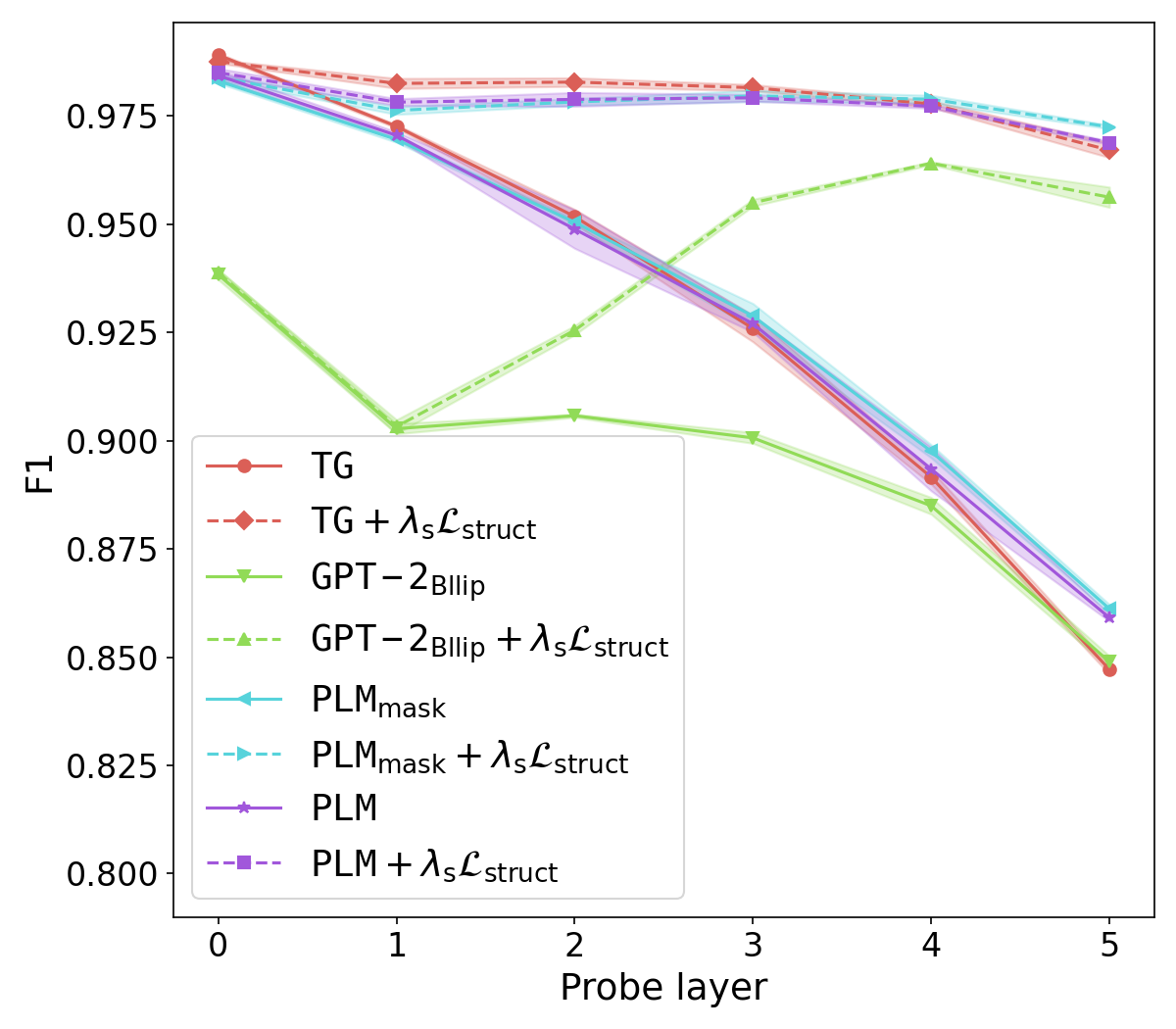}
    \caption{
    \small Constituency tree probe results for $\texttt{GPT-2}_\text{Bllip}$, $\texttt{TG}$, \texttt{PLM} and \texttt{PLM}\textsubscript{mask} on USCOCO.
    Layer 0 corresponds to text encoder output (with model-dependent dimensions), layer 1 to the 256-dimensional embedding after linear projection, layer 2 to the output of the 1\textsuperscript{st} encoder layer, etc.
    }
    \label{fig:probe}
\end{figure}

The structural loss helps to almost perfectly retain the constituency structure.
The loss matches the output (in our case visual objects) to constituency tree positions, and as the probe shows, to do so, it propagates the constituency tree information present in the text through the model.  
For \texttt{GPT-2}\textsubscript{Bllip}, except for an initial drop caused by the linear projection to a lower dimension, the loss improves probing F1 in later layers, even beyond the F1 for raw text encoder output. 
That this increase does not lead to improved layout predictions could be explained by the relevant syntax being encoded in a different, more implicit form that is harder for downstream models to learn to use.
}

\subsubsection{Computational cost}
\label{sec:cost}

{The addition of structural (parenthesis and tag) tokens to explicit syntax model input causes the number of tokens $N_c + N_\text{lin}$ to be larger than the number of tokens $N_c$ that implicit models use to encode the same sentence. 
}
\texttt{GPT-2} needs only 11 tokens on average per sentence in the COCO validation set, versus \texttt{TG} that needs 38 and \texttt{PLM} that needs 30.
This translates in a greater computational cost for the explicit syntax models.

Nevertheless, the small \texttt{TG} + $\mathcal{L}_\text{struct}$, pretrained only on BLLIP\textsubscript{LG}, outperforms the large \texttt{GPT-2-lg} that has been pretrained on a much larger dataset.
This entails multiple computational advantages: smaller memory footprint and fewer resources and less time needed for pretraining.

\section{Limitations \& future work}
One limitation of layout decoding with explicitly structured language models is the reliance upon a syntax parsing model to obtain the constituency trees for input captions. 
While syntax parsing models have shown very high performance (the parser of \citet{kitaev2018constituency} obtains an F1 score of 95.13 on the Penn Treebank \citep{marcus1993penn}, which contains longer and more syntactically complex sentences than typical COCO captions),
grammatical errors in the used prompts might result in incorrect parses and hence in worse layout generations, compared to language models without explicit syntax (that do not need a parser). 
We leave an investigation into this phenomenon for further research. 
However, we do note an increased performance of layout generation even with only trivial right-branching trees over implicit syntax models (visible in Table \ref{tab:results-final}), which might be an indication of robustness against grammatical errors for models that explicitly encode syntax.

Furthermore, while we show that explicitly modelling syntax improves layout prediction for absurd situations, this out-of-distribution generation task still remains difficult even for the 
best layout predictor models: there is a 35-37\% drop in F1 score and 17-26\% drop in $\Fpw$ on USCOCO compared to the in-domain test set. 
The introduction of USCOCO allows further research to evaluate new layout generation models on their out-of-distribution and absurd generation capabilities.

Very recent work has prompted the GPT-4 API to generate SVG or TikZ code that can be rendered into schematic images, which can then be used to guide the generation of more detailed images \citep{bubeck2023sparks,zhang2023controllable}.
The layout prediction models discussed in our paper generate bounding box coordinates and class labels, which are hard to directly compare to code or rendered images.
Moreover, we studied the role that explicit grammar can play for robustness with respect to absurd inputs, which would not have been possible with the GPT-4 API.
However, using LLMs for layout prediction can be a promising direction for future work.

\section{Conclusion}
We evaluated models that implicitly and explicitly capture the syntax of a sentence and assessed how well they retain this syntax in the representations when performing the downstream task of layout prediction of objects on a 2D canvas.  
To test compositional understanding, we collected a test set of grammatically correct sentences and layouts describing compositions of entities and relations that unlikely have been seen during training. We introduced a novel parallel decoder for layout prediction based on a transformer architecture, but most importantly we proposed a novel contrastive structural loss that enforces the encoding of syntax structure in the representation of a visual scene and show that it increases generalization to unexpected compositions resulting in large performance gains in the task of 2D spatial layout prediction conditioned on text. The loss has the potential to be used in other generation tasks that condition on structured input, which could be investigated in future work. Our research is a step forward in retaining structured knowledge in neural models.

\section*{Acknowledgements}
This work is part of the CALCULUS project, which is
funded by the ERC Advanced Grant H2020-ERC-2017
ADG 788506.\footnote{\url{https://calculus-project.eu/}} 
It also received funding from the Research Foundation -- Flanders (FWO) under Grant Agreement No. G078618N.
The computational resources and services used in this work were provided by the VSC (Flemish Supercomputer Center), funded by 
FWO and the Flemish Government.
We thank the reviewers and action editors of the TACL journal for their insightful feedback and comments.

\iftoggle{showold}{%
We proposed a parallel layout decoding model that is more accurate and efficient than previous models.
We introduced a new evaluation dataset USCOCO that tests the generalization to unseen and unexpected situations of sentence-to-layout models. 
We have shown that the performance of current state-of-the-art models is very poor on this dataset. 
However, explicitly representing syntax in the neural models and pushing the models to retain this information with a newly introduced structural loss proved to be beneficial. 
The results demonstrate a clear improvement of 
performance on the task of sentence-to-layout prediction, especially when dealing with unexpected situations. 
Our work shows the importance of retaining structured knowledge in neural models.
}{}

\iftoggle{showold}{%
\begin{figure*}
    \centering
    \includegraphics[width=0.8\textwidth]{figures/abs_test_abs_sent_micro_tag_macro_f1_size.png}
    \caption{Constituency tree probe results for $\texttt{GPT-2}$ on USCOCO. Layer 0 corresponds to before projecting to a lower dimension, layer 1 to right after the projection.}
    \label{fig:probe-2}
\end{figure*}
}{}

\bibliography{main}
\bibliographystyle{acl_natbib}

\clearpage
\appendix

\section{Structural loss $\mathcal{L}_\text{struct}$}
\label{app:struct-loss}

The structural loss $\mathcal{L}_\text{struct}(m) = \mathcal{L}_\text{struct}(\{v_k\}_k)$ for a sample $m$ takes as input the following.
\begin{itemize}
    \item Of sample $m$; the set of visual object embeddings $v^m_k$ that the layout predictor computes (and from which labels and positions are predicted).
    \item Of sample $m$: the set of tree positional embeddings $\{e_j^{m}\}_{0 \leq j < 2N_c -1}$ for both leaf and parent nodes of the constituency tree, computed according to \citet{shiv2019novel}. We omit the extra superscript as in $e^\text{pos}$ here for clarity, and to better distinguish between embeddings for sample $m$ and for other samples $n$, but $e_j^{m}$ here denotes the same tree positional embedding $e_j^\text{pos}$ as in the main text.
    \item Of other samples $n$ in the same minibatch: the set of visual object embeddings $v_k^n$ for those samples.
    \item Of other samples $n$ in the same minibatch: set of tree positional embeddings $\{e_j^{n}\}_{0 \leq j < 2N_c -1}$ for those samples.
\end{itemize}

The loss maximizes the probability that visual object embeddings $V_m = \{v_k^m\}_k$ for sample $m$ are matched with tree positional embeddings $E_m = \{e_j^{m}\}_j$ for sample $m$ and not those of other samples $n$, and vice versa.
\begin{align}
\mathcal{L}_\text{struct}(m) = - \log P(V_m|E_m) %
- \log P(E_m|V_m)
\end{align}
The probabilities are computed as a softmax over similarity scores $S(m,n)$ between samples in the minibatch (batch size $M$).
\begin{align}
P(V_m|E_n) &= \frac{\exp \left(\gamma_{1} S\left(E_n, V_m\right)\right)}{\sum_{i=1}^{M} \exp \left(\gamma_{1} S\left(E_n, V_i\right)\right)} \\
P(E_m|V_n) &= \frac{\exp \left(\gamma_{1} S\left(E_m, V_n\right)\right)}{\sum_{i=1}^{M} \exp \left(\gamma_{1} S\left(E_i, V_n\right)\right)}
\end{align}
These probabilities only differ in what is being summed over in the denominator, the similarity scores $S(m,n)$ are being computed identically.

The similarity score for 2 samples $m,n$ is computed as a sum of scores $s(v^m_{i}, c^n_{i})$, each of which quantifies how well a particular visual embedding $v^m_i$ of sample $m$ aligns with the set of tree positional embeddings $\{e^n_j\}_j$ of sample $n$, where the $e^n_j$ are expressed through a context vector $c^n_i$. The layout predictor predicts $N_\text{obj}$ visual objects for sample $m$.
\begin{align}
    S(E_n, V_m)=\log \left(\sum_{i=1}^{N_\text{obj}} \exp \left(\gamma_{2} s\left(v^m_{i}, c^n_{i}\right)\right)\right)^{\frac{1}{\gamma_{2}}}
\end{align}
Similarity function $s$ returns the cosine similarity between vectors $v^m_{i}$ and $c^n_{i}$. 

The context vector $c^n_i$ is computed with attention, with tree positional embeddings $\{e^n_j\}_j$ of sample $n$ as values and visual embedding $v^m_i$ of sample $m$ as query.
\begin{align}
    c^n_i & =\sum_{j=1}^{2N_c -1} \alpha_{i,j} e^n_j \\
    \alpha_{i,j} &= \frac{\exp \left(\gamma_{3} r(v^m_i, e^n_j)\right)}{\sum_{l=1}^{2N_c -1} \exp \left(\gamma_{3} r(v^m_i, e^n_l)\right)} 
\end{align}

Alignment scores $r(v^m_i, e^n_j)$ quantify how well a particular visual embedding $v^m_i$ of sample $m$ aligns with a particular tree positional embedding $e^n_j$ of sample $n$.
\begin{align}
    r(v^m_i, e^n_j)=\frac{\exp \left(s(v^m_i, e^n_j)\right)}{\sum_{k=1}^{N_\text{obj}} \exp \left(s(v^m_k, e^n_j)\right)}
\end{align}
Here, $s$ returns the cosine similarity again. The alignment scores are normalized so that the alignment scores for one tree positional embedding $e^n_j$ with all visual embeddings $\{v^m_k\}_k$ sum to 1.
Constants $\gamma_1, \gamma_2, \gamma_3$ are hyperparameters.

\section{USCOCO dataset}
\label{app:uscoco}

The USCOCO dataset is created in three steps:

\begin{enumerate}
    \item Firstly we obtained co-reference links between bounding boxes, of the images of the COCO2017 validation split, and their image caption parts. These annotations where retrieved with annotators from Amazon Mechanical Turk (AMT). This AMT task is visualized in figure %
    \ref{fig:mturk_coref_ex}.
    
    \item Then the caption parts and bounding boxes were automatically replaced to create potential samples of the USCOCO dataset. For this replacement a random COCO object class was sampled out of all the other classes that were not part of the original class' superclass and while excluding the superclasses summarized in Table \ref{tab:exl_sup_classes}. The text part is replaced by the new sampled class name taking in account the preposition and whether or not the word is in singular or plural. The bounding boxes are then replaced in 4 different manners:
    
    \begin{table}
    \centering
    \small
    \begin{tabular}{lc}
    \toprule
    Super class & Excluded superclasses \\
    \midrule
    vehicle & outdoor \\
    outdoor & vehicle \\
    kitchen & food, indoor \\
    food & kitchen, indoor \\
    indoor & kitchen, food \\
    furniture & appliance \\
    appliance & furniture\\
    \bottomrule
    \end{tabular}
    \caption{Excluded COCO superclasses for the generation of the USCOCO dataset.}
    \label{tab:exl_sup_classes}
\end{table}

\begin{figure*}
\begin{center}
\includegraphics[width=0.8\textwidth]{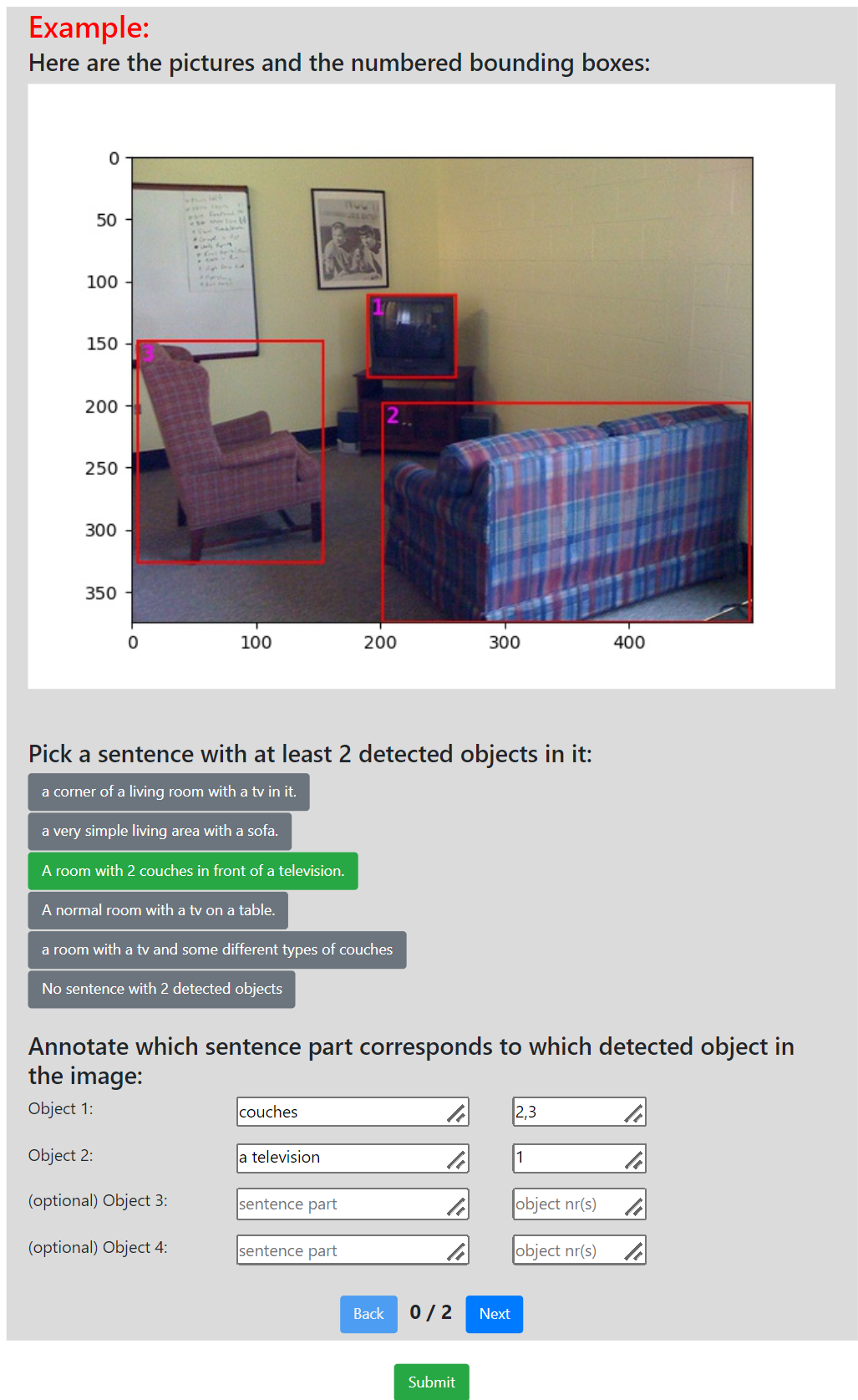}
\caption{Example of the AMT assignment for the co-reference retrieval.}
\label{fig:mturk_coref_ex}
\end{center}
\end{figure*}

\begin{figure*}
\begin{center}
\includegraphics[width=0.8\textwidth]{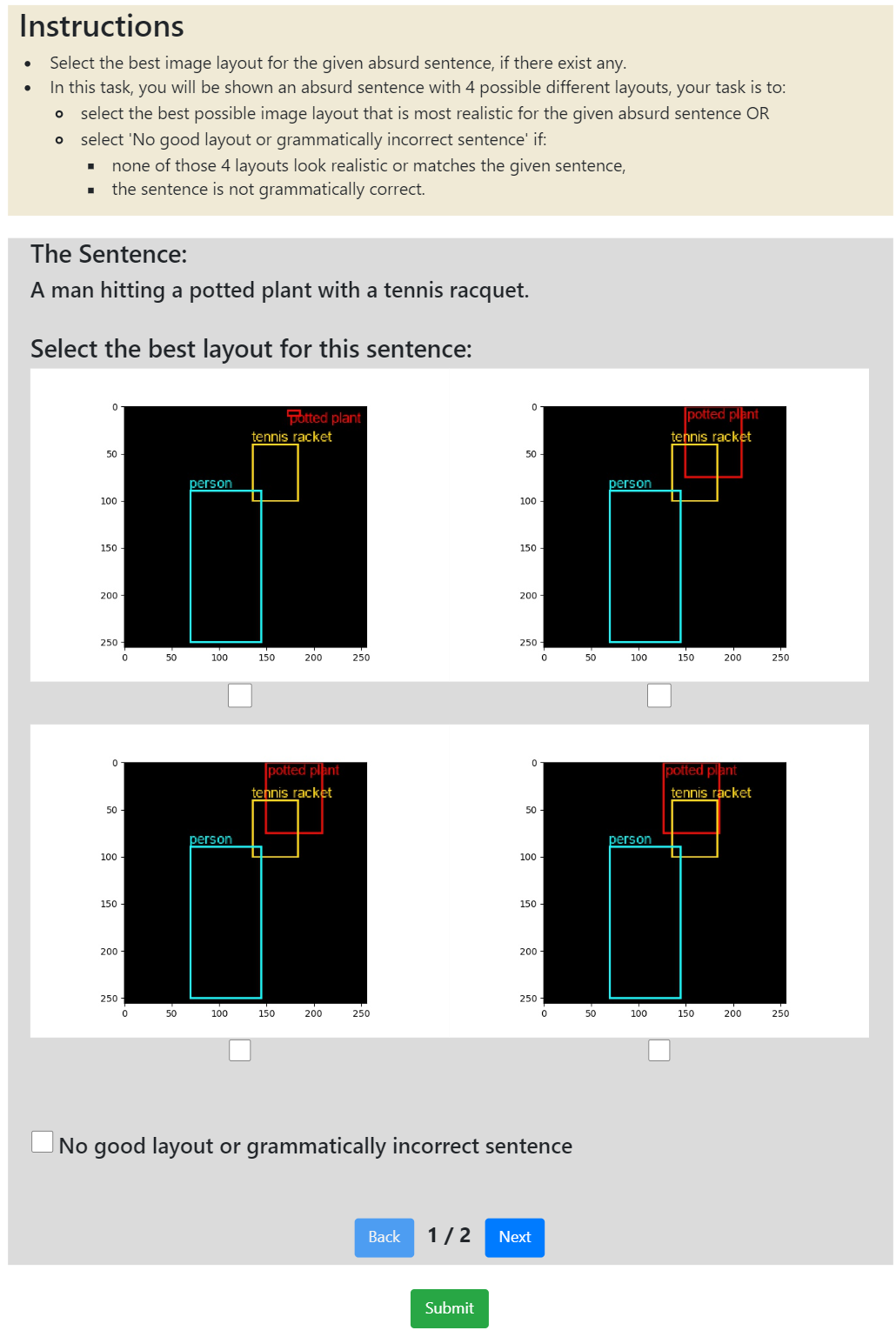}
\caption{AMT assignment for the layout selection.}
\label{fig:mturk_layout}
\end{center}
\end{figure*}

    \begin{enumerate}
        \item The original bounding box locations are kept the same.
        \item The average object size of the new class is scaled by the size of the closest object divided by the average size of the corresponding class. This new bounding box is positioned with the same middle point as the original box.
        \item The average object size of the new class is scaled by the size of the closest object divided by the average size of the corresponding class. This new bounding box is positioned with the same x-as distance to its closest object.
        \item The average object size of the new class is scaled by the size of the closest object divided by the average size of the corresponding class. This new bounding box is positioned with the same y-as distance to its closest object.
    \end{enumerate}
    The script for this automatic replacement will be made available on the project's github repo\footnote{\url{https://github.com/rubencart/USCOCO}}.
    
    \item In the last step the obtained captions of unusual situations and their layouts are verified with AMT. The annotators were asked to verify that the sentence is grammatically correct and if it is to select the best possible layout or none if they were all badly generated (this task is visualized in figure \ref{fig:mturk_layout}). Each sample of step 2 was verified by 3 different annotators. Samples where at least 2 annotators agreed on the same layout and none selected that the sentence was not grammatically correct where added to the final USCOCO dataset.
    
\end{enumerate}

From the 5000 images of the COCO2017 validation set, 2682 images retrieved co-references with at least 2 annotated sentence parts. We replaced each sentence part of the previous step to obtain, for these 2682 images, 5413 possible USCOCO samples. After the last verification step the process resulted in 2441 verified USCOCO samples generated from 1719 different images. \\

\iftoggle{showold}{%
\begin{table*}
    \centering
    \scriptsize
\setlength\tabcolsep{3pt}
\begin{tabular}{lllllllllll}
\toprule
 {} & \multicolumn{10}{c}{$\mathcal{D}_\text{USCOCO}$} \\
    & $\lambda_\text{1}$
     & $\Fp\uparrow$ & $\Pp\uparrow$ & $\Rp\uparrow$
     & F1 $\uparrow$ & Pr $\uparrow$ & Re $\uparrow$
     & $\Fpw \uparrow$ & $\Prpw \uparrow$ & $\Repw \uparrow$ 
     \\
\midrule
\texttt{GPT-2} &  & .179 {\tiny $\pm$ .008} & .197 {\tiny $\pm$ .019} & .164 {\tiny $\pm$ .002} & .566 {\tiny $\pm$ .02} & .62 {\tiny $\pm$ .046} & .522 {\tiny $\pm$ .025} & .207 {\tiny $\pm$ .019} & .248 {\tiny $\pm$ .042} & .18 {\tiny $\pm$ .023} \\
\texttt{GPT-2} + $\lambda_\text{s} \mathcal{L}_\text{struct}$ & .25 & .184 {\tiny $\pm$ .007} & .215 {\tiny $\pm$ .017} & .161 {\tiny $\pm$ .002} & .555 {\tiny $\pm$ .011} & .652 {\tiny $\pm$ .038} & .483 {\tiny $\pm$ .008} & .187 {\tiny $\pm$ .006} & .278 {\tiny $\pm$ .027} & .142 {\tiny $\pm$ .009} \\
\texttt{GPT-2} + $\lambda_\text{s} \mathcal{L}_\text{struct}$ & .50 & .181 {\tiny $\pm$ .007} & .223 {\tiny $\pm$ .032} & .154 {\tiny $\pm$ .007} & .528 {\tiny $\pm$ .024} & .654 {\tiny $\pm$ .055} & .446 {\tiny $\pm$ .048} & .155 {\tiny $\pm$ .04} & .286 {\tiny $\pm$ .054} & .112 {\tiny $\pm$ .043} \\
\texttt{GPT-2} + $\lambda_\text{s} \mathcal{L}_\text{struct}$ & 1 & .183 {\tiny $\pm$ .005} & .223 {\tiny $\pm$ .025} & .157 {\tiny $\pm$ .005} & .504 {\tiny $\pm$ .01} & .629 {\tiny $\pm$ .045} & .422 {\tiny $\pm$ .03} & .134 {\tiny $\pm$ .024} & .258 {\tiny $\pm$ .031} & .093 {\tiny $\pm$ .025} \\
\midrule
\texttt{GPT-2-md} &  & .175 {\tiny $\pm$ .008} & .191 {\tiny $\pm$ .013} & .161 {\tiny $\pm$ .006} & .529 {\tiny $\pm$ .034} & .584 {\tiny $\pm$ .037} & .484 {\tiny $\pm$ .038} & .166 {\tiny $\pm$ .035} & .21 {\tiny $\pm$ .032} & .138 {\tiny $\pm$ .036} \\
\texttt{GPT-2-md} + $\lambda_\text{s} \mathcal{L}_\text{struct}$ & .25 & .18 {\tiny $\pm$ .018} & .214 {\tiny $\pm$ .037} & .156 {\tiny $\pm$ .013} & .516 {\tiny $\pm$ .06} & .615 {\tiny $\pm$ .102} & .446 {\tiny $\pm$ .053} & .152 {\tiny $\pm$ .039} & .253 {\tiny $\pm$ .084} & .112 {\tiny $\pm$ .035} \\
\texttt{GPT-2-md} + $\lambda_\text{s} \mathcal{L}_\text{struct}$ & .50 & .174 {\tiny $\pm$ .006} & .198 {\tiny $\pm$ .001} & .154 {\tiny $\pm$ .009} & .508 {\tiny $\pm$ .024} & .583 {\tiny $\pm$ .006} & .45 {\tiny $\pm$ .034} & .148 {\tiny $\pm$ .02} & .211 {\tiny $\pm$ .004} & .115 {\tiny $\pm$ .025} \\
\texttt{GPT-2-md} + $\lambda_\text{s} \mathcal{L}_\text{struct}$ & 1 & .155 {\tiny $\pm$ .003} & .173 {\tiny $\pm$ .008} & .141 {\tiny $\pm$ .001} & .435 {\tiny $\pm$ .014} & .496 {\tiny $\pm$ .014} & .389 {\tiny $\pm$ .023} & .097 {\tiny $\pm$ .009} & .151 {\tiny $\pm$ .025} & .072 {\tiny $\pm$ .012} \\
\midrule
\texttt{GPT-2-lg} &  & .188 {\tiny $\pm$ .005} & .188 {\tiny $\pm$ .004} & .188 {\tiny $\pm$ .005} & .61 {\tiny $\pm$ .03} & .599 {\tiny $\pm$ .027} & .622 {\tiny $\pm$ .034} & .283 {\tiny $\pm$ .047} & .271 {\tiny $\pm$ .046} & .297 {\tiny $\pm$ .048} \\
\texttt{GPT-2-lg} + $\lambda_\text{s} \mathcal{L}_\text{struct}$ & .25 & .204 {\tiny $\pm$ .007} & .226 {\tiny $\pm$ .016} & .186 {\tiny $\pm$ .004} & .59 {\tiny $\pm$ .024} & .654 {\tiny $\pm$ .007} & .539 {\tiny $\pm$ .043} & .247 {\tiny $\pm$ .039} & .313 {\tiny $\pm$ .014} & .206 {\tiny $\pm$ .048} \\
\texttt{GPT-2-lg} + $\lambda_\text{s} \mathcal{L}_\text{struct}$ & .50 & .205 {\tiny $\pm$ .009} & .231 {\tiny $\pm$ .012} & .184 {\tiny $\pm$ .008} & .628 {\tiny $\pm$ .016} & .693 {\tiny $\pm$ .007} & .574 {\tiny $\pm$ .032} & .292 {\tiny $\pm$ .025} & .366 {\tiny $\pm$ .002} & .244 {\tiny $\pm$ .036} \\
\texttt{GPT-2-lg} + $\lambda_\text{s} \mathcal{L}_\text{struct}$ & 1 & .201 {\tiny $\pm$ .005} & .221 {\tiny $\pm$ .006} & .184 {\tiny $\pm$ .005} & .579 {\tiny $\pm$ .026} & .641 {\tiny $\pm$ .027} & .528 {\tiny $\pm$ .029} & .24 {\tiny $\pm$ .035} & .306 {\tiny $\pm$ .039} & .197 {\tiny $\pm$ .033} \\
\midrule
\texttt{GPT-2}\textsubscript{Bllip} &  & .18 {\tiny $\pm$ .001} & .188 {\tiny $\pm$ .001} & .172 {\tiny $\pm$ .002} & .576 {\tiny $\pm$ .026} & .599 {\tiny $\pm$ .025} & .555 {\tiny $\pm$ .028} & .229 {\tiny $\pm$ .036} & .242 {\tiny $\pm$ .043} & .217 {\tiny $\pm$ .031} \\
\texttt{GPT-2}\textsubscript{Bllip} + $\lambda_\text{s} \mathcal{L}_\text{struct}$ & .25 & .193 {\tiny $\pm$ .002} & .213 {\tiny $\pm$ .01} & .177 {\tiny $\pm$ .005} & .567 {\tiny $\pm$ .02} & .629 {\tiny $\pm$ .023} & .517 {\tiny $\pm$ .032} & .226 {\tiny $\pm$ .025} & .287 {\tiny $\pm$ .025} & .187 {\tiny $\pm$ .032} \\
\texttt{GPT-2}\textsubscript{Bllip} + $\lambda_\text{s} \mathcal{L}_\text{struct}$ & .50 & .192 {\tiny $\pm$ .003} & .22 {\tiny $\pm$ .006} & .17 {\tiny $\pm$ .001} & .574 {\tiny $\pm$ .014} & .65 {\tiny $\pm$ .021} & .513 {\tiny $\pm$ .011} & .233 {\tiny $\pm$ .014} & .309 {\tiny $\pm$ .028} & .187 {\tiny $\pm$ .009} \\
\texttt{GPT-2}\textsubscript{Bllip} + $\lambda_\text{s} \mathcal{L}_\text{struct}$ & 1 & .189 {\tiny $\pm$ .005} & .233 {\tiny $\pm$ .01} & .159 {\tiny $\pm$ .003} & .525 {\tiny $\pm$ .014} & .66 {\tiny $\pm$ .011} & .436 {\tiny $\pm$ .016} & .17 {\tiny $\pm$ .017} & .308 {\tiny $\pm$ .02} & .118 {\tiny $\pm$ .015} \\
\midrule
\texttt{GPT-2-md}\textsubscript{Bllip} &  & .179 {\tiny $\pm$ .001} & .19 {\tiny $\pm$ .006} & .169 {\tiny $\pm$ .002} & .586 {\tiny $\pm$ .017} & .612 {\tiny $\pm$ .026} & .562 {\tiny $\pm$ .015} & .235 {\tiny $\pm$ .024} & .252 {\tiny $\pm$ .036} & .221 {\tiny $\pm$ .02} \\
\texttt{GPT-2-md}\textsubscript{Bllip} + $\lambda_\text{s} \mathcal{L}_\text{struct}$ & .25 & .189 {\tiny $\pm$ .003} & .213 {\tiny $\pm$ .009} & .169 {\tiny $\pm$ .002} & .581 {\tiny $\pm$ .031} & .658 {\tiny $\pm$ .016} & .52 {\tiny $\pm$ .041} & .239 {\tiny $\pm$ .041} & .319 {\tiny $\pm$ .028} & .192 {\tiny $\pm$ .043} \\
\texttt{GPT-2-md}\textsubscript{Bllip} + $\lambda_\text{s} \mathcal{L}_\text{struct}$ & .50 & .192 {\tiny $\pm$ .008} & .216 {\tiny $\pm$ .013} & .173 {\tiny $\pm$ .005} & .577 {\tiny $\pm$ .023} & .652 {\tiny $\pm$ .032} & .518 {\tiny $\pm$ .019} & .237 {\tiny $\pm$ .026} & .315 {\tiny $\pm$ .041} & .191 {\tiny $\pm$ .021} \\
\texttt{GPT-2-md}\textsubscript{Bllip} + $\lambda_\text{s} \mathcal{L}_\text{struct}$ & 1 & .185 {\tiny $\pm$ .006} & .21 {\tiny $\pm$ .008} & .166 {\tiny $\pm$ .01} & .539 {\tiny $\pm$ .032} & .622 {\tiny $\pm$ .033} & .476 {\tiny $\pm$ .038} & .197 {\tiny $\pm$ .038} & .282 {\tiny $\pm$ .043} & .153 {\tiny $\pm$ .037} \\
\midrule
\texttt{GPT-2-lg}\textsubscript{Bllip} &  & .183 {\tiny $\pm$ .002} & .198 {\tiny $\pm$ .003} & .17 {\tiny $\pm$ .003} & .586 {\tiny $\pm$ .019} & .622 {\tiny $\pm$ .017} & .555 {\tiny $\pm$ .021} & .233 {\tiny $\pm$ .027} & .255 {\tiny $\pm$ .032} & .215 {\tiny $\pm$ .024} \\
\texttt{GPT-2-lg}\textsubscript{Bllip} + $\lambda_\text{s} \mathcal{L}_\text{struct}$ & .25 & .194 {\tiny $\pm$ .007} & .234 {\tiny $\pm$ .01} & .166 {\tiny $\pm$ .007} & .563 {\tiny $\pm$ .008} & .673 {\tiny $\pm$ .021} & .484 {\tiny $\pm$ .003} & .2 {\tiny $\pm$ .004} & .306 {\tiny $\pm$ .022} & .149 {\tiny $\pm$ .005} \\
\texttt{GPT-2-lg}\textsubscript{Bllip} + $\lambda_\text{s} \mathcal{L}_\text{struct}$ & .50 & .196 {\tiny $\pm$ .005} & .226 {\tiny $\pm$ .015} & .173 {\tiny $\pm$ .005} & .579 {\tiny $\pm$ .006} & .665 {\tiny $\pm$ .023} & .514 {\tiny $\pm$ .02} & .234 {\tiny $\pm$ .019} & .319 {\tiny $\pm$ .005} & .186 {\tiny $\pm$ .025} \\
\texttt{GPT-2-lg}\textsubscript{Bllip} + $\lambda_\text{s} \mathcal{L}_\text{struct}$ & 1 & .19 {\tiny $\pm$ .005} & .215 {\tiny $\pm$ .004} & .17 {\tiny $\pm$ .011} & .551 {\tiny $\pm$ .035} & .636 {\tiny $\pm$ .018} & .486 {\tiny $\pm$ .047} & .205 {\tiny $\pm$ .044} & .294 {\tiny $\pm$ .027} & .159 {\tiny $\pm$ .047} \\
\midrule
\texttt{PLM} &  & .18 {\tiny $\pm$ .002} & .189 {\tiny $\pm$ .005} & .172 {\tiny $\pm$ .0} & .579 {\tiny $\pm$ .002} & .592 {\tiny $\pm$ .007} & .566 {\tiny $\pm$ .009} & .226 {\tiny $\pm$ .006} & .228 {\tiny $\pm$ .004} & .224 {\tiny $\pm$ .008} \\
\texttt{PLM} + $\lambda_\text{s} \mathcal{L}_\text{struct}$ & .25 & .192 {\tiny $\pm$ .002} & .201 {\tiny $\pm$ .009} & .185 {\tiny $\pm$ .009} & .61 {\tiny $\pm$ .033} & .629 {\tiny $\pm$ .003} & .594 {\tiny $\pm$ .062} & .282 {\tiny $\pm$ .048} & .299 {\tiny $\pm$ .011} & .271 {\tiny $\pm$ .074} \\
\texttt{PLM} + $\lambda_\text{s} \mathcal{L}_\text{struct}$ & .50 & .192 {\tiny $\pm$ .006} & .21 {\tiny $\pm$ .024} & .177 {\tiny $\pm$ .006} & .602 {\tiny $\pm$ .019} & .649 {\tiny $\pm$ .039} & .565 {\tiny $\pm$ .045} & .267 {\tiny $\pm$ .029} & .313 {\tiny $\pm$ .025} & .238 {\tiny $\pm$ .055} \\
\texttt{PLM} + $\lambda_\text{s} \mathcal{L}_\text{struct}$ & 1 & .194 {\tiny $\pm$ .004} & .207 {\tiny $\pm$ .01} & .182 {\tiny $\pm$ .003} & .6 {\tiny $\pm$ .021} & .646 {\tiny $\pm$ .012} & .561 {\tiny $\pm$ .04} & .269 {\tiny $\pm$ .032} & .319 {\tiny $\pm$ .009} & .235 {\tiny $\pm$ .047} \\
\midrule
\texttt{PLM}\textsubscript{mask} &  & .176 {\tiny $\pm$ .005} & .184 {\tiny $\pm$ .008} & .168 {\tiny $\pm$ .002} & .588 {\tiny $\pm$ .01} & .603 {\tiny $\pm$ .011} & .574 {\tiny $\pm$ .017} & .234 {\tiny $\pm$ .012} & .236 {\tiny $\pm$ .012} & .232 {\tiny $\pm$ .019} \\
\texttt{PLM}\textsubscript{mask} + $\lambda_\text{s} \mathcal{L}_\text{struct}$ & .25 & .191 {\tiny $\pm$ .007} & .202 {\tiny $\pm$ .011} & .181 {\tiny $\pm$ .006} & .612 {\tiny $\pm$ .024} & .633 {\tiny $\pm$ .031} & .592 {\tiny $\pm$ .022} & .28 {\tiny $\pm$ .039} & .297 {\tiny $\pm$ .044} & .266 {\tiny $\pm$ .036} \\
\texttt{PLM}\textsubscript{mask} + $\lambda_\text{s} \mathcal{L}_\text{struct}$ & .50 & .186 {\tiny $\pm$ .005} & .2 {\tiny $\pm$ .017} & .175 {\tiny $\pm$ .006} & .609 {\tiny $\pm$ .013} & .643 {\tiny $\pm$ .031} & .582 {\tiny $\pm$ .052} & .28 {\tiny $\pm$ .027} & .311 {\tiny $\pm$ .017} & .261 {\tiny $\pm$ .061} \\
\texttt{PLM}\textsubscript{mask} + $\lambda_\text{s} \mathcal{L}_\text{struct}$ & 1 & .19 {\tiny $\pm$ .005} & .203 {\tiny $\pm$ .01} & .179 {\tiny $\pm$ .002} & .599 {\tiny $\pm$ .018} & .645 {\tiny $\pm$ .003} & .56 {\tiny $\pm$ .029} & .272 {\tiny $\pm$ .029} & .319 {\tiny $\pm$ .015} & .237 {\tiny $\pm$ .037} \\
\midrule
\texttt{TG} &  & .185 {\tiny $\pm$ .004} & .192 {\tiny $\pm$ .006} & .178 {\tiny $\pm$ .005} & .6 {\tiny $\pm$ .004} & .612 {\tiny $\pm$ .014} & .589 {\tiny $\pm$ .008} & .255 {\tiny $\pm$ .005} & .259 {\tiny $\pm$ .004} & .251 {\tiny $\pm$ .012} \\
\texttt{TG} + $\lambda_\text{s} \mathcal{L}_\text{struct}$ & .25 & .192 {\tiny $\pm$ .008} & .206 {\tiny $\pm$ .016} & .18 {\tiny $\pm$ .002} & .641 {\tiny $\pm$ .018} & .673 {\tiny $\pm$ .032} & .612 {\tiny $\pm$ .025} & .318 {\tiny $\pm$ .026} & .351 {\tiny $\pm$ .036} & .292 {\tiny $\pm$ .033} \\
\texttt{TG} + $\lambda_\text{s} \mathcal{L}_\text{struct}$ & .50 & .199 {\tiny $\pm$ .011} & .218 {\tiny $\pm$ .016} & .183 {\tiny $\pm$ .01} & .621 {\tiny $\pm$ .013} & .672 {\tiny $\pm$ .04} & .578 {\tiny $\pm$ .011} & .293 {\tiny $\pm$ .009} & .35 {\tiny $\pm$ .041} & .254 {\tiny $\pm$ .012} \\
\texttt{TG} + $\lambda_\text{s} \mathcal{L}_\text{struct}$ & 1 & .196 {\tiny $\pm$ .006} & .22 {\tiny $\pm$ .007} & .178 {\tiny $\pm$ .005} & .618 {\tiny $\pm$ .009} & .684 {\tiny $\pm$ .02} & .564 {\tiny $\pm$ .002} & .288 {\tiny $\pm$ .008} & .357 {\tiny $\pm$ .025} & .242 {\tiny $\pm$ .002} \\
\midrule
\texttt{TG-md} &  & .187 {\tiny $\pm$ .002} & .193 {\tiny $\pm$ .0} & .181 {\tiny $\pm$ .004} & .607 {\tiny $\pm$ .025} & .62 {\tiny $\pm$ .023} & .595 {\tiny $\pm$ .027} & .267 {\tiny $\pm$ .029} & .275 {\tiny $\pm$ .027} & .26 {\tiny $\pm$ .031} \\
\texttt{TG-md} + $\lambda_\text{s} \mathcal{L}_\text{struct}$ & .25 & .191 {\tiny $\pm$ .008} & .198 {\tiny $\pm$ .01} & .185 {\tiny $\pm$ .006} & .641 {\tiny $\pm$ .02} & .65 {\tiny $\pm$ .013} & .632 {\tiny $\pm$ .027} & .321 {\tiny $\pm$ .031} & .326 {\tiny $\pm$ .031} & .315 {\tiny $\pm$ .033} \\
\texttt{TG-md} + $\lambda_\text{s} \mathcal{L}_\text{struct}$ & .50 & .197 {\tiny $\pm$ .001} & .21 {\tiny $\pm$ .004} & .186 {\tiny $\pm$ .002} & .636 {\tiny $\pm$ .013} & .668 {\tiny $\pm$ .015} & .608 {\tiny $\pm$ .022} & .316 {\tiny $\pm$ .018} & .35 {\tiny $\pm$ .01} & .29 {\tiny $\pm$ .028} \\
\texttt{TG-md} + $\lambda_\text{s} \mathcal{L}_\text{struct}$ & 1 & .192 {\tiny $\pm$ .004} & .213 {\tiny $\pm$ .004} & .175 {\tiny $\pm$ .008} & .614 {\tiny $\pm$ .017} & .665 {\tiny $\pm$ .028} & .571 {\tiny $\pm$ .019} & .287 {\tiny $\pm$ .021} & .342 {\tiny $\pm$ .033} & .249 {\tiny $\pm$ .021} \\
\midrule
\texttt{TG-lg} &  & .183 {\tiny $\pm$ .008} & .189 {\tiny $\pm$ .011} & .178 {\tiny $\pm$ .007} & .621 {\tiny $\pm$ .014} & .622 {\tiny $\pm$ .025} & .619 {\tiny $\pm$ .003} & .283 {\tiny $\pm$ .017} & .278 {\tiny $\pm$ .029} & .288 {\tiny $\pm$ .005} \\
\texttt{TG-lg} + $\lambda_\text{s} \mathcal{L}_\text{struct}$ & .25 & .195 {\tiny $\pm$ .006} & .204 {\tiny $\pm$ .012} & .187 {\tiny $\pm$ .005} & .645 {\tiny $\pm$ .01} & .661 {\tiny $\pm$ .014} & .63 {\tiny $\pm$ .023} & .327 {\tiny $\pm$ .018} & .343 {\tiny $\pm$ .015} & .313 {\tiny $\pm$ .028} \\
\texttt{TG-lg} + $\lambda_\text{s} \mathcal{L}_\text{struct}$ & .50 & .197 {\tiny $\pm$ .002} & .21 {\tiny $\pm$ .01} & .185 {\tiny $\pm$ .004} & .63 {\tiny $\pm$ .023} & .67 {\tiny $\pm$ .016} & .596 {\tiny $\pm$ .042} & .304 {\tiny $\pm$ .034} & .348 {\tiny $\pm$ .023} & .271 {\tiny $\pm$ .048} \\
\texttt{TG-lg} + $\lambda_\text{s} \mathcal{L}_\text{struct}$ & 1 & .188 {\tiny $\pm$ .008} & .198 {\tiny $\pm$ .011} & .179 {\tiny $\pm$ .007} & .613 {\tiny $\pm$ .021} & .635 {\tiny $\pm$ .03} & .593 {\tiny $\pm$ .021} & .294 {\tiny $\pm$ .029} & .318 {\tiny $\pm$ .04} & .274 {\tiny $\pm$ .026} \\
\midrule
\texttt{TG}\textsubscript{RB} &  & .178 {\tiny $\pm$ .004} & .184 {\tiny $\pm$ .001} & .173 {\tiny $\pm$ .007} & .571 {\tiny $\pm$ .011} & .577 {\tiny $\pm$ .009} & .565 {\tiny $\pm$ .023} & .216 {\tiny $\pm$ .016} & .211 {\tiny $\pm$ .018} & .223 {\tiny $\pm$ .028} \\
\texttt{TG}\textsubscript{RB} + $\lambda_\text{s} \mathcal{L}_\text{struct}$ & .25 & .186 {\tiny $\pm$ .005} & .194 {\tiny $\pm$ .006} & .179 {\tiny $\pm$ .004} & .6 {\tiny $\pm$ .028} & .613 {\tiny $\pm$ .032} & .588 {\tiny $\pm$ .024} & .266 {\tiny $\pm$ .041} & .273 {\tiny $\pm$ .053} & .26 {\tiny $\pm$ .033} \\
\texttt{TG}\textsubscript{RB} + $\lambda_\text{s} \mathcal{L}_\text{struct}$ & .50 & .189 {\tiny $\pm$ .007} & .203 {\tiny $\pm$ .009} & .178 {\tiny $\pm$ .006} & .606 {\tiny $\pm$ .017} & .647 {\tiny $\pm$ .033} & .57 {\tiny $\pm$ .011} & .278 {\tiny $\pm$ .023} & .322 {\tiny $\pm$ .043} & .245 {\tiny $\pm$ .013} \\
\texttt{TG}\textsubscript{RB} + $\lambda_\text{s} \mathcal{L}_\text{struct}$ & 1 & .192 {\tiny $\pm$ .005} & .2 {\tiny $\pm$ .009} & .184 {\tiny $\pm$ .002} & .583 {\tiny $\pm$ .009} & .609 {\tiny $\pm$ .022} & .56 {\tiny $\pm$ .013} & .26 {\tiny $\pm$ .014} & .288 {\tiny $\pm$ .025} & .239 {\tiny $\pm$ .017} \\
\midrule
\texttt{CLIP-T2} &  & .185 {\tiny $\pm$ .002} & .184 {\tiny $\pm$ .006} & .186 {\tiny $\pm$ .002} & .668 {\tiny $\pm$ .007} & .647 {\tiny $\pm$ .01} & .689 {\tiny $\pm$ .007} & .356 {\tiny $\pm$ .008} & .329 {\tiny $\pm$ .008} & .387 {\tiny $\pm$ .01} \\
\texttt{CLIP-T2} + $\lambda_\text{s} \mathcal{L}_\text{struct}$ & .25 & .201 {\tiny $\pm$ .003} & .218 {\tiny $\pm$ .006} & .186 {\tiny $\pm$ .005} & .644 {\tiny $\pm$ .023} & .705 {\tiny $\pm$ .007} & .593 {\tiny $\pm$ .036} & .315 {\tiny $\pm$ .035} & .386 {\tiny $\pm$ .019} & .267 {\tiny $\pm$ .041} \\
\texttt{CLIP-T2} + $\lambda_\text{s} \mathcal{L}_\text{struct}$ & .50 & .203 {\tiny $\pm$ .013} & .222 {\tiny $\pm$ .014} & .188 {\tiny $\pm$ .011} & .646 {\tiny $\pm$ .018} & .719 {\tiny $\pm$ .01} & .587 {\tiny $\pm$ .024} & .317 {\tiny $\pm$ .027} & .407 {\tiny $\pm$ .02} & .26 {\tiny $\pm$ .029} \\
\texttt{CLIP-T2} + $\lambda_\text{s} \mathcal{L}_\text{struct}$ & 1 & .199 {\tiny $\pm$ .005} & .233 {\tiny $\pm$ .021} & .174 {\tiny $\pm$ .005} & .594 {\tiny $\pm$ .029} & .703 {\tiny $\pm$ .051} & .516 {\tiny $\pm$ .034} & .245 {\tiny $\pm$ .04} & .371 {\tiny $\pm$ .066} & .184 {\tiny $\pm$ .035} \\
\midrule
\texttt{LLama-7B} &  & .179 {\tiny $\pm$ .007} & .183 {\tiny $\pm$ .01} & .174 {\tiny $\pm$ .005} & .583 {\tiny $\pm$ .011} & .589 {\tiny $\pm$ .019} & .576 {\tiny $\pm$ .014} & .231 {\tiny $\pm$ .014} & .234 {\tiny $\pm$ .017} & .228 {\tiny $\pm$ .017} \\
\texttt{LLama-7B} + $\lambda_\text{s} \mathcal{L}_\text{struct}$ & .25 & .192 {\tiny $\pm$ .01} & .202 {\tiny $\pm$ .008} & .183 {\tiny $\pm$ .011} & .602 {\tiny $\pm$ .02} & .63 {\tiny $\pm$ .026} & .576 {\tiny $\pm$ .017} & .26 {\tiny $\pm$ .026} & .287 {\tiny $\pm$ .033} & .238 {\tiny $\pm$ .023} \\
\texttt{LLama-7B} + $\lambda_\text{s} \mathcal{L}_\text{struct}$ & .50 & .192 {\tiny $\pm$ .006} & .203 {\tiny $\pm$ .007} & .181 {\tiny $\pm$ .005} & .577 {\tiny $\pm$ .011} & .61 {\tiny $\pm$ .005} & .547 {\tiny $\pm$ .017} & .229 {\tiny $\pm$ .013} & .256 {\tiny $\pm$ .002} & .208 {\tiny $\pm$ .021} \\
\texttt{LLama-7B} + $\lambda_\text{s} \mathcal{L}_\text{struct}$ & 1 & .191 {\tiny $\pm$ .004} & .206 {\tiny $\pm$ .009} & .178 {\tiny $\pm$ .001} & .553 {\tiny $\pm$ .008} & .597 {\tiny $\pm$ .022} & .515 {\tiny $\pm$ .009} & .198 {\tiny $\pm$ .008} & .238 {\tiny $\pm$ .022} & .171 {\tiny $\pm$ .007} \\
\texttt{LLama-33B} &  & .181 {\tiny $\pm$ .006} & .184 {\tiny $\pm$ .005} & .177 {\tiny $\pm$ .009} & .577 {\tiny $\pm$ .008} & .577 {\tiny $\pm$ .006} & .577 {\tiny $\pm$ .019} & .225 {\tiny $\pm$ .011} & .221 {\tiny $\pm$ .005} & .229 {\tiny $\pm$ .025} \\
\bottomrule
\end{tabular}

    \caption{
    Text encoders with implicit (\texttt{GPT-2}\textsubscript{Bllip}, \texttt{GPT-2}, \texttt{RoBERTa}, \texttt{LLama}, \texttt{CLIP}) and explicit (\texttt{PLM}, \texttt{PLM}\textsubscript{mask}, \texttt{TG}) syntax, and structural loss results on USCOCO: F1, precision and recall, with and without IoU threshold and pairwise distance weighted. All entries use the \texttt{PAR} layout predictor. 
    \texttt{TG}\textsubscript{RB} is the \texttt{TG} model but using trivial right-branching trees (during both pretraining and finetuning).
    \texttt{GPT-2}, \texttt{GPT-2}\textsubscript{Bllip}, \texttt{PLM}, \texttt{PLM}\textsubscript{mask} and \texttt{TG} use the smallest \texttt{GPT-2} architecture ($125$M params). \texttt{GPT-2-md}, \texttt{GPT-2-md}\textsubscript{Bllip} and \texttt{TG-md} use the medium-sized \texttt{GPT-2} architecture ($350$M params), and \texttt{GPT-2-lg}, \texttt{GPT-2-lg}\textsubscript{Bllip} and \texttt{TG-lg} use the larger \texttt{GPT-2} architecture ($750$M params). 
    \texttt{RoBERTa} and \texttt{RoBERTa-lg} use the base ($125$M params) and large ($350$M params) \texttt{RoBERTa} architecture, respectively.
    \texttt{LLama} is the large model from Facebook with 7B to 33B parameters.
    \texttt{CLIP-T2} are the token embeddings from the second to last layer of the OpenAI CLIP model (150M params)
    }
    \label{tab:results-qian-old}
\end{table*}
}{}

\end{document}